\documentclass{article}

\usepackage[eandd, final]{neurips_2026}
\usepackage[utf8]{inputenc}
\usepackage[T1]{fontenc}
\usepackage{hyperref}
\usepackage{url}
\usepackage{booktabs}
\usepackage{amsfonts}
\usepackage{amsmath}
\usepackage{amssymb}
\usepackage{microtype}
\usepackage{xcolor}
\usepackage{graphicx}
\usepackage{tabularx}
\usepackage{array}
\usepackage{makecell}
\usepackage{threeparttable}
\usepackage{wrapfig}
\usepackage{float}
\usepackage{listings}

\usepackage{subcaption}
\usepackage{enumitem}
\usepackage{multirow}
\usepackage{pifont}

\setcitestyle{numbers}

\usepackage{amsmath,amsfonts,bm}

\def\eqref#1{equation~\ref{#1}}
\def\1{\bm{1}}

\DeclareMathAlphabet{\mathsfit}{\encodingdefault}{\sfdefault}{m}{sl}
\SetMathAlphabet{\mathsfit}{bold}{\encodingdefault}{\sfdefault}{bx}{n}

\definecolor{color1}{HTML}{1F77B4} %
\definecolor{color2}{HTML}{E377C2} %
\definecolor{color3}{HTML}{2CA02C} %
\definecolor{color4}{HTML}{D62728} %
\definecolor{color5}{HTML}{9467BD} %
\definecolor{color6}{HTML}{8C564B} %

\definecolor{mplblue}{HTML}{1F77B4}
\definecolor{mplorange}{HTML}{FF7F0E}
\definecolor{mplgreen}{HTML}{2CA02C}
\definecolor{mplred}{HTML}{D62728}

\title{A Matched-Budget Audit Framework for Recaptioned Image-Text Supervision Distributions}

\author{%
  Giyeong Oh%\textsuperscript{\ding{169}}
  \And
  Junghun Park%\textsuperscript{\ding{170}}
  \And
  Yuhan Bae%\textsuperscript{$\spadesuit$}
  \And
  Youngjae Yu%\textsuperscript{$\spadesuit$}
}

\newcommand{\afflegend}{%
\begin{center}
{
Seoul National University \\
\texttt{\{hard2251,impri,yuhanbae06,youngjaeyu\}@snu.ac.kr}
}
\end{center}
}

\newcommand{\githubicon}{\raisebox{-0.15em}{\includegraphics[height=1em]{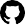}}}
\newcommand{\hficon}{\raisebox{-0.15em}{\includegraphics[height=1.1em]{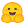}}}

\newcommand{\linkline}{%
\begin{center}
{\normalfont\footnotesize
\githubicon\,\href{https://github.com/BootsofLagrangian/audit-recap-t2i}{\texttt{GitHub}}
\quad
\hficon\,\href{https://huggingface.co/collections/BootsofLagrangian/recaptioned-image-text}{\texttt{Hugging Face}}
}
\end{center}
}

\begin{document}

\maketitle
\vspace{-2.5em}
\afflegend
\vspace{-1.2em}
\linkline

% \begingroup
% \renewcommand{\thefootnote}{}
% \footnotetext{%
% \begin{tabular}{@{}l@{\hspace{0.25em}}l@{}}
% \textsuperscript{\ding{169}} & Interdisciplinary Program in Artificial Intelligence \\
% \textsuperscript{\ding{170}} & Department of Computer Science and Engineering \\
% \textsuperscript{$\spadesuit$} & School of Transdisciplinary Innovations
% \end{tabular}%
% }
% \endgroup

\begin{abstract}
Recaptioned image-text corpora are now standard for text-to-image (T2I) training, with vision--language model (VLM) captioners replacing sparse alt-text by dense descriptions.
A recaptioned corpus is a supervision distribution induced by a documented captioning policy ($\pi$), captioner ($V_c$), and source corpus ($C$).
Length-correlated proxies miss caption-register artifacts and downstream T2I benchmarks entangle the corpus with training choices, so this distribution is hard to audit at corpus scale.
We introduce a reusable matched-budget audit framework for recaptioned supervision distributions $D_{\pi,V_c,C}$: at a fixed text budget of $B = 64$ it reports a five-axis profile spanning prompt-side coverage, image-conditioned faithfulness, and caption-surface health, with claimed controllable basic units (CBU) as the common claim unit.
We instantiate the framework on seven paired comparisons over five public source corpora.
Across the four cross-corpus pairs, the released surface raises supported CBU per caption by $+3.39$ to $+6.36$ under both Qwen and Gemma Judges, and on CC12M the same framework exposes a long-vs-dense frontier that is consistent across both judges and four budgets.
We release the audited multi-source recap corpus ($\approx$ 490M) together with the audit-artifact bundle.
\end{abstract}

\section{Introduction}
\label{sec:intro}

Text-to-image (T2I) models~\citep{betker2024dalle3, rombach2022sd, podell2024sdxl, esser2024sd3, labs2024flux, qwen2025qwenimage, hunyuanimage2025} train on caption-image pairs and respond to user prompts; captions describe images and prompts state user intent.
DALL-E 3 popularized dense VLM captions for T2I training~\citep{betker2024dalle3}, and open releases apply the recipe at corpus scale~\citep{segalis2023recap, li2024recapdatacomp, singla2024pixelprose, chen2023sharegpt4v}.
VLM captioners inherit \emph{describe-style frames} such as third-person openers (e.g., \texttt{The image shows\dots}) and broad-frame templates (\texttt{This is a [adjective]\dots}), caption-only scaffolds that sit outside the register of user prompts~\citep{hirota2024descriptive}.
Structured-caption methods bridge the train-inference gap with prompt-style schemas~\citep{merchant2025structured}, and their enumerated scaffolds (e.g., \texttt{1.\ \dots~2.\ \dots}) form a further describe-style frame.
Every recaptioning policy injects bias: the policy determines \textit{which biases enter the corpus} and how much of the caption text carries controllable content.

Captions can be audited at corpus level before any T2I training: caption-only and image-conditioned audit tools already exist~\citep{friedman2023vendi, hirota2024descriptive, rohrbach2018object, hu2023tifa, cho2024dsg, petryk2024aloha, hessel2021clipscore, longclip2024}, and public recap-corpus releases rarely apply them jointly to the corpus as a supervision distribution.
In practice releases are read by length~\citep{li2024recapdatacomp, yu2024capsfusion} or by downstream T2I benchmarks that entangle the captioning corpus with augmentation policy and testset register~\citep{hu2023tifa, cho2024dsg, ghosh2023geneval, huang2025t2i, betker2024dalle3, brack2025train}, so the captioning step is the natural audit point.
We therefore treat each recaptioned corpus as a documented audit target $D_{\pi, V_c, C}$, the joint over caption $c$ and source image $x$ induced by captioner $V_c$ under policy $\pi$ on source corpus $C$, and introduce a matched-budget audit framework for comparing such targets.
The framework needs caption text, source-row alignment, and image access for the faithfulness axis; at a fixed text budget $B$ it decomposes prompt-side coverage, image-conditioned faithfulness, and caption-surface health into five reported axes.

Our policy injects the register of user prompts by one simple rule: write the caption as a user writes a prompt, from the dominant subject through scene structure to camera framing (Appendix~\ref{app:vlm_templates}).
We release the resulting multi-source recap corpus as one such audit target and pair its caption set, a \emph{surface}, against public reference surfaces.
We instantiate the audit on seven paired comparisons over five public source corpora (DataComp~\citep{gadre2023datacomp}, CC12M~\citep{changpinyo2021cc12m}, LAION-pop~\citep{schuhmann2022laion5b}, PD12M~\citep{meyer2024pd12m}, Danbooru~\citep{danbooru2023}); four release-only families (CommonCatalog~\citep{gokaslan2024commoncanvas}, LAION-Aesthetics~\citep{schuhmann2022laion5b}, LAION-HighRes-Aesthetic~\citep{schuhmann2022laion5b}, Megalith-CC0~\citep{pdExtendedHF}) appear in the release artifact.
The audit returns a per-axis profile: across the four cross-corpus pairs, the released surface raises claimed controllable basic units (CBU) per caption by $+2.91$ to $+6.14$, and both Qwen and Gemma Judges report higher supported CBU per caption ($+3.39$ to $+6.36$) and lower unsupported risk ($-0.046$ to $-0.159$); on CC12M the same framework exposes a long-vs-dense frontier consistent under both Judges and across budgets $B \in \{16, 32, 48, 64\}$.
With the captioner fixed, this policy alone raises supported CBU per caption by $+3.0$ to $+3.6$ over a plain detailed-caption instruction (Section~\ref{sec:captioner_control}).

\begin{figure}[!t]
\centering

\begin{subfigure}[c]{0.53\linewidth}
    \centering
    \includegraphics[width=\linewidth]{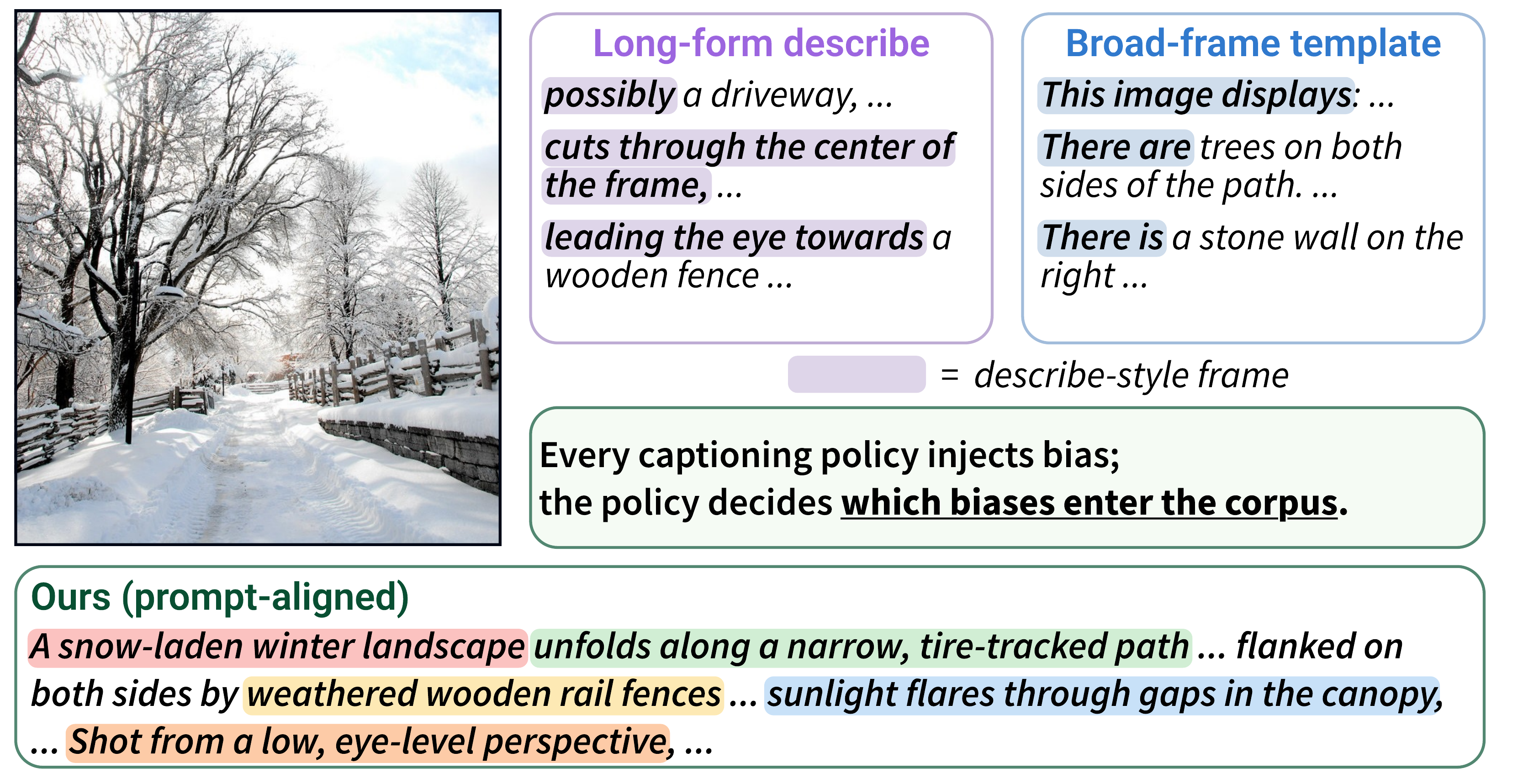}
    \caption{Policy sets the caption register.}
    \label{fig:teaser_opener}
\end{subfigure}
\hfill
\begin{subfigure}[c]{0.45\linewidth}
    \centering
    \includegraphics[width=\linewidth]{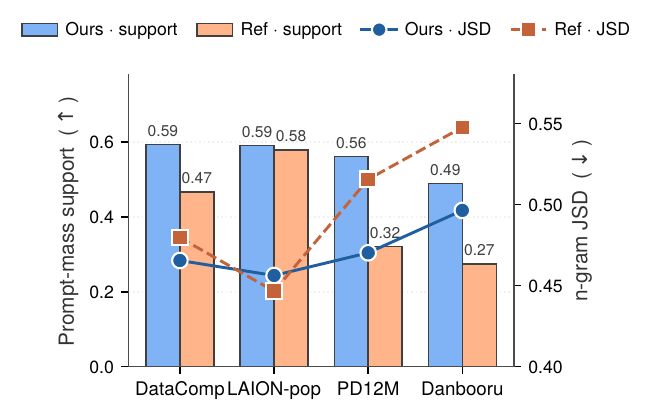}
    \caption{Prompt-pool diagnostics.}
    \label{fig:teaser_pool_support}
\end{subfigure}

\caption{\textbf{VLM bias is inevitable; policy decides which biases survive.}
\textit{Left:}\protect\footnotemark{} describe-style frame prefix concentration drops on the released surface.
\textit{Right:} averaged across 7 prompt pools per cross-corpus pair, Ours raises prompt-mass support on every pair and lowers $n$-gram Jensen-Shannon Divergence (JSD) against the pool distribution on DataComp, PD12M, and Danbooru, with the length-matched LAION-pop pair within $0.01$; bars on the left axis report support, markers on the right axis report JSD; definitions are in Section~\ref{sec:protocol:axes}.}
\label{fig:teaser}
\end{figure}
\footnotetext{Photo: ``Toronto: Riverdale farm'' by The City of Toronto (Flickr), CC BY 2.0, cropped.}

We make five contributions:
\begin{itemize}[leftmargin=1.2em]
\item \textbf{Audit target.} We formalize recaptioned image-text data as supervision distributions $D_{\pi, V_c, C}$ induced by a documented policy, captioner, and source corpus, separating the audit target from any downstream T2I model trained on it.
\item \textbf{Audit framework.} We define a corpus-agnostic audit operator with a fixed text budget and five axes of prompt-side coverage, image-conditioned faithfulness, and caption-surface health.
\item \textbf{CBU bridge.} We introduce controllable basic units (CBUs) as claim-level units counted caption-only and re-asked against the paired image, letting claimed density, supported yield, and unsupported risk share one claim set.
\item \textbf{Paired comparisons.} We instantiate the framework on describe-style frames (Section~\ref{sec:prelim:gap}), four cross-corpus pairs (Section~\ref{sec:cross}), a CC12M long-vs-dense frontier (Section~\ref{sec:cc12m}), and a same-captioner policy ablation locating the cross-corpus gap on the policy axis (Section~\ref{sec:captioner_control}).
\item \textbf{Corpus release.} We release the audited multi-source corpus of captions for $\approx $490M image identities, unique within each source family, together with the audit-artifact bundle (lexical tokenizer, prompt pools, request templates, CBU schema, judge schema, caption hashes, Croissant core fields, Responsible-AI metadata~\citep{akhtar2024croissant, jain2024croissantrai, gebru2021datasheets}) so the same profile can be re-run on future $(\pi, V_c, C)$ audit targets.
\end{itemize}

\section{Related Work}
\label{sec:related}

\paragraph{VLM recaptioning.}
DALL-E 3 first showed that dense VLM captions sharply improve T2I prompt-following~\citep{betker2024dalle3}.
Open follow-ups scaled the recipe across released corpora: Recap-DataComp-1B~\citep{li2024recapdatacomp}, PixelProse~\citep{singla2024pixelprose}, and ShareGPT4V~\citep{chen2023sharegpt4v}.
Principled prompt design and structured caption schemas extend the lineage~\citep{segalis2023recap, merchant2025structured}.
Releases vary captioner, policy, and length together, so a comparison across releases reads all three at once.

\paragraph{Prompt-side adaptation and the caption-prompt gap.}
Inference-side adaptation rewrites~\citep{hao2023optimizing, zhan2024capr}, expands~\citep{datta2024prompt}, or optimizes~\citep{manas2024opt2i,yeh2026tipo} user prompts toward the captioner-style training distribution; public prompt corpora and community repositories for Stable Diffusion, SDXL, and FLUX seed our prompt-reference pools (Appendix~\ref{app:protocol}).
On the caption side, length and claim density are separate quantities~\citep{detailness2025, urbanek2024dci}, and most public recap pipelines write in describe-style frames~\citep{betker2024dalle3, li2024recapdatacomp, singla2024pixelprose, hirota2024descriptive}; Re-LAION-Caption~\citep{merchant2025structured} narrows the gap with a four-part instruction template whose enumeration is its own surface form, so prompt coverage follows from the register a policy selects.

\paragraph{Image-grounded faithfulness.}
Visual question answering (VQA) decomposition checks caption faithfulness claim by claim against the image~\citep{rohrbach2018object, hu2023tifa, cho2024dsg, petryk2024aloha}.
CLIPScore~\citep{hessel2021clipscore} reports global image-text alignment, and encoder truncation~\citep{longclip2024} limits its reach on long captions.
Image-output benchmarks~\citep{ghosh2023geneval, huang2025t2i, hu2024ella} measure the compositional faithfulness of generated images, and recap-corpus releases rarely carry the matching caption-only and image-caption audits.

\paragraph{Corpus-level auditability.}
Prior work supplies the probes: caption length, descriptive-bias measures~\citep{hirota2024descriptive}, contrastive image-text scores~\citep{hessel2021clipscore, longclip2024}, VQA decomposition~\citep{hu2023tifa, cho2024dsg, petryk2024aloha}, diversity estimates~\citep{friedman2023vendi, naeem2020prdc}, and downstream T2I benchmarks~\citep{ghosh2023geneval, huang2025t2i}.
A third-party re-audit needs source rows, the captioner and its policy, a text budget, prompt-reference pools, a claim schema, a judge schema, and an artifact manifest, and our framework assembles this set.
Table~\ref{tab:prior_audits} places it among the closest dataset and caption studies and claim-level metrics: it carries claim-level image verification to the corpus level and adds a matched text budget and the register of user prompts.

\begin{table}[t]
\centering
\caption{\textbf{Dataset and caption studies (top) and claim-level metrics (middle).} \emph{Natural}: text on non-generated images; \emph{Corpus}: dataset-level analysis; \emph{Unit}: text scored per claim or object mention; \emph{Image}: text verified against the image; \emph{Budget}: fixed text window; \emph{Prompt}: register of user prompts.}
\label{tab:prior_audits}
\footnotesize
\renewcommand{\arraystretch}{1.0}
\setlength{\tabcolsep}{5.0pt}
\begin{tabular}{@{}l l cccccc@{}}
\toprule
\textbf{Work} & \textbf{Target} & \textbf{Natural} & \textbf{Corpus} & \textbf{Unit} & \textbf{Image} & \textbf{Budget} & \textbf{Prompt} \\
\midrule
REVISE~\citep{wang2022revise} & visual datasets & \checkmark & \checkmark & -- & -- & -- & -- \\
LAION's Den~\citep{birhane2023laionden} & image--alt-text pairs & \checkmark & \checkmark & -- & -- & -- & -- \\
Hirota et al.~\citep{hirota2024descriptive} & caption enrichment & \checkmark & \checkmark & \checkmark & \checkmark & -- & -- \\
Brack et al.~\citep{brack2025train} & training captions & \checkmark & \checkmark & -- & -- & -- & -- \\
\midrule
TIFA / DSG~\citep{hu2023tifa, cho2024dsg} & generated images & -- & -- & \checkmark & \checkmark & -- & -- \\
FAITHSCORE~\citep{jing2024faithscore} & VLM answers & \checkmark & -- & \checkmark & \checkmark & -- & -- \\
DCScore~\citep{ye2025dcscore} & detailed captions & \checkmark & -- & \checkmark & \checkmark & -- & -- \\
\midrule
Ours & recaptioned corpora & \checkmark & \checkmark & \checkmark & \checkmark & \checkmark & \checkmark \\
\bottomrule
\end{tabular}
\end{table}

\begin{table}[t]
\centering
\caption{\textbf{Nine recaptioned source families: five with paired audits, four release-only.}
Ours scale counts unique images per family; a dash marks a release-only family.
The dagger ${}^{\dagger}$ marks short tag-style surfaces (under $20$ lexical units).}
\label{tab:target_families}

\footnotesize
\renewcommand{\arraystretch}{1.05}
\setlength{\tabcolsep}{3.0pt}

\begin{tabularx}{\linewidth}{@{}l l r >{\raggedright\arraybackslash}X@{}}
\toprule
\textbf{Source family} 
& \textbf{Original supervision} 
& \textbf{Ours scale} 
& \textbf{Paired reference surface(s)} \\
\midrule
\multicolumn{4}{@{}l}{\emph{Photorealistic / web}} \\

DataComp~\citep{gadre2023datacomp}
& web image--text pairs
& $\approx 325.5$M
& Recap-DataComp~\citep{li2024recapdatacomp} \\

CC12M~\citep{changpinyo2021cc12m}
& web alt-text
& $\approx 11.5$M
& CC12M-LLaVA-NeXT~\citep{conceptualcaptionscc12mllavanext},
  PixelProse~\citep{singla2024pixelprose},
  CC12M-Qwen3-VL${}^{\dagger}$~\citep{cc12mQwen3VL8BHF} \\

LAION-pop~\citep{schuhmann2022laion5b}
& web alt-text
& $\approx 0.4$M
& LAION-pop-Llama~\citep{laionPopLlama32HF} \\

PD12M~\citep{meyer2024pd12m}
& Florence-2 + metadata
& $\approx 12.4$M
& PD12M released~\citep{pd12mFullHF} \\

CommonCatalog~\citep{gokaslan2024commoncanvas}
& BLIP-2 captions
& $\approx 14.6$M
& --- \\

LAION-Aesthetics~\citep{schuhmann2022laion5b}
& web alt-text
& $\approx 23.7$M
& --- \\

LAION-HighRes-Aesthetic~\citep{schuhmann2022laion5b}
& web alt-text
& $\approx 82.7$M
& \leavevmode--- \\

Megalith-CC0~\citep{pdExtendedHF}
& Florence-2 captions
& $\approx 8.1$M
& \leavevmode--- \\

\midrule
\multicolumn{4}{@{}l}{\emph{Anime / booru}} \\

Danbooru~\citep{danbooru2023}
& booru tags + metadata
& $\approx 11.3$M
& Danbooru-Florence~\citep{xiao2024florence2, danbooruFlorence2HF} \\

\bottomrule
\end{tabularx}
\end{table}

\section{Audit Targets and the Diagnostic Gap}
\label{sec:prelim}

\subsection{Notation}
\label{sec:prelim:notation}

Let $\mathcal{X}, \mathcal{T}$ denote image and text spaces, $C \subset \mathcal{X}$ a source image corpus.
A captioner $V_c$ run under policy $\pi$ on $C$ induces a paired supervision distribution
\[
  D \;=\; D_{\pi, V_c, C} \;=\; \{(c, x) : x \in C,\; c \sim V_c(x; \pi)\},
\] 
with joint $D_{cx} = D$ and caption marginal $D_c$.
We write $D^{\star} = D_{\pi^{\star}, V_c^{\star}, C}$ for \emph{Ours}.
A T2I model $G_\theta : \mathcal{T} \to \mathcal{X}$ trained on $D$ generates $\hat{x} = G_\theta(p)$ at inference from a user prompt $p$.

VLMs fill three roles: the captioner $V_c$ writes training text, the claim extractor $\phi$ reads claims from a caption, and the \emph{Judge} $V_J$ re-asks each claim against the paired image.
A \emph{controllable basic unit} (CBU) is a caption claim in one of eight semantic visual-claim types (Section~\ref{sec:protocol:axes}); $\phi$ returns the CBUs in a caption, a subset of the claim space $\mathcal{C}$, and the Judge labels each (CBU, image) pair:
\[
  \phi : \mathcal{T} \to 2^{\mathcal{C}},
  \qquad
  V_J : \mathcal{C} \times \mathcal{X} \to \{\mathrm{supported},\,\mathrm{unsupported},\,\mathrm{uncertain}\}.
\]
Here $V_c = $ \texttt{Qwen3.5-35B-A3B-FP8}~\citep{qwen35}; $\phi$ and the \emph{Qwen Judge} share \texttt{Qwen3.5-397B-A17B-FP8}; the \emph{Gemma Judge} is \texttt{Gemma-4-31B-IT}~\citep{gemma4}.
Both Judges are open-weight VLMs served locally with deterministic, schema-constrained answers across all eight claim types. The Qwen Judge is the $397$B-parameter checkpoint of the captioner's family and also serves as the extractor; the Gemma Judge comes from an independently trained family and re-asks the same extracted claims, so agreement between the two Judges is agreement across model families.
Checkpoint URLs and request templates are in Appendix~\ref{app:vlm_templates}, and Appendix~\ref{app:glossary} lists the notation and audit vocabulary.
The documented distribution $D_{\pi, V_c, C}$ is the \emph{audit target}, and each caption set over the same source rows is a \emph{surface}; the \emph{audit framework} of Section~\ref{sec:protocol} audits $D^{\star}$ and the reference surfaces.

\subsection{Target Source Families and Audited Surfaces}
\label{sec:prelim:targets}
The release covers nine public source families, five of them with paired audits (Table~\ref{tab:target_families}).
For each family the released \emph{Ours} surface is produced under a single versioned, domain-branched policy on $V_c$ that writes in the order of a user prompt, opening with the dominant subject and expanding through scene-level structure (Appendix~\ref{app:vlm_templates}); reference surfaces caption the same source rows under other captioners and policies.
Each comparison fixes $C$ and varies $(V_c, \pi)$, whereas a single downstream T2I score folds captioner, policy, and training configuration into one number.

\subsection{The Diagnostic Gap}
\label{sec:prelim:gap}

Public recap releases are typically read by mean caption length and downstream T2I score.
Four inexpensive text-only diagnostics separate the audited surfaces (Table~\ref{tab:phenomenon_meta_opener}; up to $1\mathrm{M}$ raw captions per surface, Appendix~\ref{app:phenomenon_method} for full operationalization):
\begin{itemize}[leftmargin=1.2em,itemsep=1pt,topsep=2pt]
\item \textbf{Opener rate.} Share of captions whose leading sentence matches a regex catalog of third-person caption-style frames (e.g., \texttt{The image [shows|depicts|\dots] \dots}); lower means fewer captions open with this frame.
\item \textbf{Top-100 raw / content prefix mass.} Share of captions whose first five lexical units fall among the $100$ most frequent such prefixes, computed raw and on the \emph{content prefix} left after leading articles and frame words are dropped; lower is less concentrated.
\item \textbf{Within-corpus distinct-3.} Unique-3-gram fraction over the corpus; higher is less repetitive.
\end{itemize}
These CPU-cost probes read caption-level markers and corpus-level distributions on any single-caption release, and releases with several captions per image also admit cross-caption consistency checks.
On several references content-prefix mass stays high and distinct-3 stays low, a repetition that caption length leaves unreported.
Brack et al.~\citep{brack2025train} ablate length, density, and persona at the downstream-T2I level with a fixed captioning instruction that \emph{suppresses} the opener (e.g., \texttt{"Do not start with `The image'"}). The multi-axis audit of Section~\ref{sec:protocol} compares the surfaces that captioning policies \emph{select}, at corpus scale and on the released text, so it also applies to pipelines whose policies are undisclosed.

\begin{table}[t]
\centering
\caption{\textbf{Public recap surfaces concentrate beyond opener boilerplate.} Phenomenon descriptors on up to $1\mathrm{M}$ raw captions per surface; Avg lex is mean lexical-unit length. Ours keeps the opener rate at $1.3$--$2.3\%$ on every source and, against every long-form reference, lowers top-100 raw and content mass and raises Distinct-3; dagger ${}^{\dagger}$ marks short tag-style surfaces ($<20$ lex), whose small 3-gram count raises Distinct-3. Per-source-family breakdown in Appendix Table~\ref{tab:phenomenon_full}.}
\label{tab:phenomenon_meta_opener}
\footnotesize
\renewcommand{\arraystretch}{1.0}
\setlength{\tabcolsep}{2pt}
\begin{tabular}{llrrrrr}
\toprule
\textbf{Source} & \textbf{Reference release} & \textbf{Avg lex} & \textbf{Opener\,$\downarrow$} & \textbf{Top-100 raw\,$\downarrow$} & \textbf{Top-100 content\,$\downarrow$} & \textbf{Distinct-3\,$\uparrow$} \\
\midrule
DataComp   & Recap-DataComp~\citep{li2024recapdatacomp}                  & $50.9$  & $0.36\%$   & $27.99\%$ & $17.53\%$ & $11.49\%$ \\
CC12M      & CC12M-LLaVA-NeXT~\citep{conceptualcaptionscc12mllavanext}    & $72.2$  & $0.24\%$   & $22.97\%$ & $12.39\%$ & $13.81\%$ \\
           & PixelProse~\citep{singla2024pixelprose}                      & $89.5$  & $100.00\%$ & $65.07\%$ & $8.25\%$  & $12.56\%$ \\
           & CC12M-Qwen3-VL${}^{\dagger}$~\citep{cc12mQwen3VL8BHF}         & $11.6$  & $0.00\%$   & $9.40\%$  & $5.87\%$  & $37.04\%$ \\
LAION-pop  & LAION-pop-Llama~\citep{laionPopLlama32HF}                   & $180.3$ & $0.78\%$   & $23.68\%$ & $14.54\%$ & $24.83\%$ \\
PD12M      & PD12M released~\citep{pd12mFullHF}                           & $39.7$  & $99.98\%$  & $90.73\%$ & $43.10\%$ & $2.40\%$  \\
Danbooru   & Danbooru-Florence~\citep{xiao2024florence2, danbooruFlorence2HF} & $43.6$ & $77.29\%$  & $99.46\%$ & $61.10\%$ & $1.34\%$  \\
\midrule
\multicolumn{2}{@{}l}{Ours (comparison range)} & $164$--$189$ & $1.27$--$2.28\%$ & $4.30$--$15.82\%$ & $2.36$--$10.34\%$ & $13.49$--$36.87\%$ \\
\bottomrule
\end{tabular}
\end{table}

\section{Matched-Budget Audit Framework}
\label{sec:protocol}

The framework is a reusable audit operator over recaptioned supervision distributions.
Given an audit target $D_{\pi, V_c, C}$ (Section~\ref{sec:prelim:notation}) at text budget $B$, prompt-reference pools sampled per paired comparison, 
a CBU extractor $\phi$, and an image-conditioned judge $V_J$, the audit returns a five-axis profile spanning text budget, prompt-pool support, claimed CBU density, surface concentration, and image-conditioned support / risk.
The audit reads the supervision signal $G_\theta$ would consume, before any generator is trained.
It runs four passes over source-aligned rows: it cuts every caption to the text budget $B$, computes deterministic text statistics on the resulting window, extracts the CBUs the window states, and re-asks each CBU against the paired image (Table~\ref{tab:axes}).
We instantiate the operator at $B = 64$ on the seven paired comparisons of Section~\ref{sec:prelim:targets}. Text statistics count the budget in lexical units (regex tokens after Unicode normalization, independent of BPE~\citep{sennrich2016bpe} and text-encoder~\citep{radford2021clip, raffel2020t5} tokenizers); the claim passes read the first $B$ whitespace-delimited words, and the lexical units inside that window normalize claim density (Appendix~\ref{app:protocol}).
$B = 64$ covers the typical length of short reference releases (PD12M-released $\approx 40$ lex, Recap-DataComp $\approx 51$ lex) and bounds per-caption VLM extraction cost.

\subsection{Caption~$\leftrightarrow$~Prompt Structural Properties}
\label{sec:protocol:desiderata}

For $G_\theta$ to be controllable at inference, $D_{cx}$ should have three complementary properties.

\paragraph{Coverage.}
$D_c$ should structurally span the prompt distribution $P$ in the T2I sense: a controllable T2I model composes unseen scenes (\texttt{an astronaut riding a horse on Mars}; \texttt{an avocado-shaped armchair}) from components (e.g., astronaut, horse, Mars-like landscape, avocado, armchair) that appear as supported claims in $D_c$.
Coverage is therefore read at the component level: prompt-pool support measures the $n$-gram mass of typical prompts that the captions carry, and claimed CBU density counts the components each caption states.

\paragraph{Faithfulness.}
An unsupported claim binds text to content absent from $x$ and gives $G_\theta$ an inconsistent training pair.
With $s(c, x) = |\{\mu \in \phi(c) : V_J(\mu, x) = \mathrm{supported}\}|$ and $u(c, x)$ analogous for unsupported, faithfulness measures the claim-weighted risk
\[
  \rho(D_{cx}) \;=\; \mathbb{E}[u(c, x)] \,/\, \mathbb{E}[|\phi(c)|].
\]
Supported yield and risk read the same $(\text{claim}, \text{image})$ pairs: yield counts the claims the Judge finds supported by the image, and risk is the share it answers \emph{no} for.

\paragraph{Surface health.}
The marginal $D_c$ should spend its prefix on visual content; describe-style frames concentrate it on shared form.
Surface health is a prefix-level text property read through prefix concentration and repetition, and Section~\ref{sec:prelim:gap} gives the corpus-scale evidence on public recap surfaces.

\subsection{Axes and Metrics}
\label{sec:protocol:axes}

The three properties decompose into five measurable axes.

\paragraph{Controllable basic units.}
A CBU is a caption claim of one of eight semantic visual-claim types (object, attribute, relation, count, style, camera, lighting, text-rendering); $\phi$ extracts CBUs caption-only, $V_J$ re-asks each CBU against the paired image (Appendix~\ref{app:vlm_templates}).
CBUs serve as the common claim set behind claimed density, supported yield, and unsupported risk.
With $c_{\le B}$ the $B$-word window of $c$ and $\ell(\cdot)$ its length in lexical units,
\begin{equation}
\label{eq:cbu_metrics}
\mathrm{CBU/cap}(D_c, B) \;=\; \mathbb{E}_{c \sim D_c}\!\left[|\phi(c_{\le B})|\right],
\quad
\mathrm{CBU/100\,lex}(D_c, B) \;=\; \frac{100 \cdot \mathbb{E}_{c}\!\left[|\phi(c_{\le B})|\right]}{\mathbb{E}_{c}\!\left[\ell(c_{\le B})\right]}.
\end{equation}
Table~\ref{tab:axes} lists the five axes, each reading $D_c$, its relation to user prompts, or $D_{cx}$.
\emph{$B$-eligibility} is the share of captions reaching $B$ lexical units; \emph{prompt-mass support} is the share of pool prompt mass whose $n$-grams appear in $D_c$, and \emph{$n$-gram JSD}~\citep{lin1991jsd} compares the two $n$-gram distributions over the seven prompt-reference pools.
Both Judges re-ask every $\mu \in \phi(c)$ against $x$ under one question schema~\citep{hu2023tifa, cho2024dsg}; a single open-vocabulary judge keeps one answer schema and one question set across the eight claim types, and the CBU schema also admits specialized OCR, detection, and counting models on the types they cover.

\begin{table}[t]
\centering
\caption{\textbf{Audit axes.} Arrows mark the better direction; \emph{Rows} is the cap on paired rows per slice.}
\label{tab:axes}
\footnotesize
\renewcommand{\arraystretch}{1.0}
\setlength{\tabcolsep}{5.0pt}
\begin{tabular}{@{}l l l l r@{}}
\toprule
\textbf{Axis} & \textbf{Property} & \textbf{Input} & \textbf{Metric} & \textbf{Rows} \\
\midrule
Text budget & Coverage & $D_c$ & Avg.\ lex, $B$-eligibility & $1$M \\
Prompt-pool support & Coverage & $D_c$, pools & prompt-mass support$\uparrow$, $n$-gram JSD$\downarrow$ & $250$k \\
Claimed density & Coverage & $D_c$ & CBU/cap$\uparrow$, CBU/100\,lex & $5$k \\
Surface concentration & Health & $D_c$ & top-100 prefix mass$\downarrow$, distinct-3$\uparrow$ & $1$M \\
Support and risk & Faithfulness & $D_{cx}$ & $\mathbb{E}[s]\uparrow$, $\rho\downarrow$ & $5$k \\
\bottomrule
\end{tabular}
\end{table}

\section{Framework Instantiation}
\label{sec:eval}

Seven paired comparisons hold source rows fixed and vary the caption surface: four cross-corpus pairs (Section~\ref{sec:cross}) and three CC12M references (Section~\ref{sec:cc12m}). A captioner control fixes the captioner and varies the policy (Section~\ref{sec:captioner_control}), human annotators verify image support on CC12M (Section~\ref{sec:human_cbu}), and encoder probes report their sensitivity (Section~\ref{sec:diagnostics}).
Text statistics use up to $1$M paired rows per slice and prompt-pool support up to $250{,}000$; the claim extractor and both Judges read $\approx 5{,}000$ of those rows per surface ($4{,}494$ on CC12M; Appendix Tables~\ref{tab:vqa_questions},~\ref{tab:cc12m_denominators}), and the two Judges answer one question set; prompt-pool support is computed against seven sampled prompt-reference pools per slice (Appendix~\ref{app:protocol}).

\subsection{Cross-Corpus Results}
\label{sec:cross}

Table~\ref{tab:cross_full} reports the four cross-corpus pairs at $B = 64$.
On every pair Ours raises supported CBU per caption and lowers unsupported risk under both Qwen and Gemma Judges; the supported-yield gain spans $+3.4$ to $+6.4$ CBU per caption and the risk drop spans $0.05$ to $0.16$ across the eight (pair, judge) cells.
Per-pool prompt-mass deltas (Appendix Figure~\ref{fig:per_pool_heatmap}) place Ours above the reference on $5$--$7$ of the seven prompt pools per slice; the Pool-wins column of Table~\ref{tab:cross_full} aggregates this reading.
The six claim types other than count and relation reproduce the ordering: Ours raises supported CBU per caption by $+2.66$ to $+5.22$ and lowers risk in all eight (pair, judge) cells (Appendix Table~\ref{tab:excl_count_relation}).

\newcommand{\crosspair}[2]{%
  \makebox[3.1em][r]{\ensuremath{#1}}%
  \ensuremath{\,\to\,}%
  \makebox[3.1em][l]{\ensuremath{#2}}%
}

\begin{table}[!t]
\centering
\caption{\textbf{Cross-corpus results at $B = 64$.} Cells $a \!\to\! b$ report Ref $\!\to\!$ Ours (bold = Ours). Pool-wins counts the seven prompt pools where Ours raises prompt-mass support; Qwen and Gemma Judges re-ask the same CBUs on $\approx 5{,}000$ paired captions per surface; Risk is unsupported / claimed CBU. Per-pool deltas in Appendix Figure~\ref{fig:per_pool_heatmap}, full descriptors in Appendix Tables~\ref{tab:vqa_questions},~\ref{tab:phenomenon_full}.}
\label{tab:cross_full}
\setlength{\tabcolsep}{1.3pt}
\scalebox{0.699}{
\begin{tabular}{@{}l c c c c c c c@{}}
\toprule
& & & & \multicolumn{2}{c}{\textbf{Qwen Judge}}
& \multicolumn{2}{c}{\textbf{Gemma Judge}} \\
\cmidrule(lr){5-6} \cmidrule(l){7-8}
\textbf{Dataset}
& \textbf{Avg lex}
& \textbf{CBU/cap$\uparrow$}
& \textbf{Pool-wins$\uparrow$}
& \textbf{Sup.\ CBU/cap$\uparrow$}
& \textbf{Risk$\downarrow$}
& \textbf{Sup.\ CBU/cap$\uparrow$}
& \textbf{Risk$\downarrow$} \\
\midrule
DataComp
& \crosspair{50.9}{175.5}
& \crosspair{10.44}{\mathbf{14.45}}
& $\mathbf{6}/7$
& \crosspair{8.49}{\mathbf{13.73}}
& \crosspair{0.177}{\mathbf{0.035}}
& \crosspair{8.00}{\mathbf{12.94}}
& \crosspair{0.219}{\mathbf{0.081}} \\
LAION-pop
& \crosspair{180.3}{182.9}
& \crosspair{11.91}{\mathbf{14.82}}
& $\mathbf{5}/7$
& \crosspair{10.80}{\mathbf{14.22}}
& \crosspair{0.077}{\mathbf{0.031}}
& \crosspair{10.22}{\mathbf{13.61}}
& \crosspair{0.113}{\mathbf{0.060}} \\
PD12M
& \crosspair{39.7}{189.1}
& \crosspair{9.78}{\mathbf{15.02}}
& $\mathbf{7}/7$
& \crosspair{8.61}{\mathbf{14.29}}
& \crosspair{0.103}{\mathbf{0.034}}
& \crosspair{8.23}{\mathbf{13.53}}
& \crosspair{0.131}{\mathbf{0.066}} \\
Danbooru
& \crosspair{43.6}{164.5}
& \crosspair{8.18}{\mathbf{14.33}}
& $\mathbf{7}/7$
& \crosspair{6.38}{\mathbf{12.74}}
& \crosspair{0.217}{\mathbf{0.058}}
& \crosspair{6.15}{\mathbf{11.85}}
& \crosspair{0.235}{\mathbf{0.094}} \\
\bottomrule
\end{tabular}}
\end{table}

\paragraph{The gain holds at matched length.}
Ours is $3$--$5\times$ longer than each reference except LAION-pop, where the two surfaces are length-matched ($180.3$ vs.\ $182.9$ lex) and Ours leads on yield, risk, and pool-mass support (Table~\ref{tab:cross_full}, LAION-pop row).
On short reference releases (PD12M-released, Recap-DataComp) CBU/100lex reads claims per retained lexical unit over a partly filled window, so Table~\ref{tab:cross_full} reports CBU/cap, the claims available inside the common budget (Eq.~\ref{eq:cbu_metrics}).

\subsection{Captioner Control: One Captioner, Two Policies}
\label{sec:captioner_control}

Cross-corpus comparisons pair Ours against released surfaces produced by different captioners.
The control fixes the captioner $V_c = $ \texttt{Qwen3.5-35B-A3B-FP8} and its decoding and swaps only the policy $\pi$: the same captioner writes a \emph{naive surface} on the same $4{,}494$ CC12M and $4{,}775$ DataComp images under the single instruction of Recap-DataComp~\citep{li2024recapdatacomp}, \texttt{``Please generate a detailed caption of this image. Please be as descriptive as possible.''}
The audit pipeline ($\phi$, $V_J$, lexical tokenizer, $B = 64$) is identical to the rest of Section~\ref{sec:eval}; full table in Appendix~\ref{app:naive_qwen35_baseline}.

\paragraph{The policy carries yield and surface health.}
On both source families the released policy raises claimed CBU per caption by $+3.7$ to $+3.9$ and supported CBU per caption by $+3.0$ to $+3.6$ under both Judges, at a risk $0.01$ to $0.02$ above the naive surface (Table~\ref{tab:captioner_control_cbu}).
On CC12M the naive surface is $1.8\times$ as long as Ours and more concentrated (Table~\ref{tab:captioner_control_surface}): frequent prefixes such as \texttt{``\dots\ richly detailed''} ($157/4{,}494$ captions) lie outside the opener regex of Section~\ref{sec:prelim:gap} and lift its content-prefix mass to $0.33$, above every released CC12M surface on the same images ($0.07$--$0.16$).
With the captioner and its decoding fixed, the prompt-ordered policy produces the yield gain on both source families.

\begin{table}[t]
\centering
\caption{\textbf{Captioner-control: same $V_c$, different $\pi$.} Both policies caption the same $\approx 4.5$--$4.8$k images per source with the same decoding; the Qwen extractor $\phi$ supplies the claims and both Judges answer the same questions (Appendix~\ref{app:naive_qwen35_baseline}).}
\label{tab:captioner_control_cbu}
\renewcommand{\arraystretch}{1.0}
\setlength{\tabcolsep}{4.0pt}
\scalebox{0.75}{
\begin{tabular}{@{}ll rr rr rr@{}}
\toprule
&
& \multicolumn{2}{c}{\textbf{Surface stats}}
& \multicolumn{2}{c}{\textbf{Qwen Judge}}
& \multicolumn{2}{c}{\textbf{Gemma Judge}} \\
\cmidrule(lr){3-4} \cmidrule(lr){5-6} \cmidrule(l){7-8}
\textbf{Dataset}
& \textbf{Surface}
& \textbf{CBU/cap}\,$\uparrow$
& \textbf{CBU/100lex}
& \textbf{Sup.\ CBU/cap}\,$\uparrow$
& \textbf{Risk}\,$\downarrow$
& \textbf{Sup.\ CBU/cap}\,$\uparrow$
& \textbf{Risk}\,$\downarrow$ \\
\midrule
\multirow{2}{*}{CC12M}
& Ours                          & $\mathbf{15.21}$ & $23.16$ & $\mathbf{14.60}$ & $0.030$ & $\mathbf{13.82}$ & $0.066$ \\
& Naive Qwen3.5-35B-A3B         & $11.32$ & $17.26$ & $11.03$ & $0.022$ & $10.61$ & $0.045$ \\
\addlinespace[0.2em]
\multirow{2}{*}{DataComp}
& Ours                          & $\mathbf{14.60}$ & $22.23$ & $\mathbf{13.90}$ & $0.036$ & $\mathbf{13.07}$ & $0.080$ \\
& Naive Qwen3.5-35B-A3B         & $10.95$ & $16.66$ & $10.56$ & $0.026$ & $10.07$ & $0.057$ \\
\bottomrule
\end{tabular}}
\end{table}

\begin{table}[t]
\centering
\caption{\textbf{Surface concentration on the CC12M Naive surface.} Section~\ref{sec:prelim:gap} diagnostics on the CC12M images of Table~\ref{tab:captioner_control_cbu}. Lex overflow$_{248}$ = share of captions over $248$ lexical units; Long-form refs.\ = range over CC12M-LLaVA-NeXT and PixelProse on the same images.}
\label{tab:captioner_control_surface}
\renewcommand{\arraystretch}{1.0}
\setlength{\tabcolsep}{4.5pt}
\scalebox{0.8}{
\begin{tabular}{@{}l ccc r@{}}
\toprule
\textbf{Metric}
& \textbf{Ours}
& \textbf{Naive}
& \textbf{Naive$-$Ours}
& \textbf{Long-form refs.} \\
\midrule
Mean lex                              & $186.6$ & $328.1$  & $+141.5$ & $67.5$--$91.4$ \\
Lex overflow$_{248}$                  & $6.9\%$ & $91.2\%$ & $+84.3$ pp & $0.1$--$0.8\%$ \\
Top-100 raw prefix mass\,$\downarrow$ & $0.14$  & $0.60$   & $+0.46$  & $0.29$--$0.71$ \\
Top-100 content prefix mass\,$\downarrow$ & $0.09$ & $0.33$ & $+0.24$  & $0.13$--$0.16$ \\
Distinct-3-gram rate\,$\uparrow$      & $0.59$  & $0.57$   & $-0.02$  & $0.40$--$0.49$ \\
\bottomrule
\end{tabular}}
\end{table}

\subsection{CC12M: A Long-vs-Dense Frontier Case Study}
\label{sec:cc12m}

CC12M carries three released comparison surfaces with distinct policies: PixelProse hard-codes openers and descriptive-narrator personas (\texttt{``Start with the words `This image displays:'\,''}, five-prompt mixture from~\citep{singla2024pixelprose}); CC12M-LLaVA-NeXT is a long-form release without explicit opener hard-coding; CC12M-Qwen3-VL${}^{\dagger}$ is a short tag-style release ($\approx 12$ lex).

\begin{figure}[t]
\centering
\includegraphics[width=0.46\linewidth]{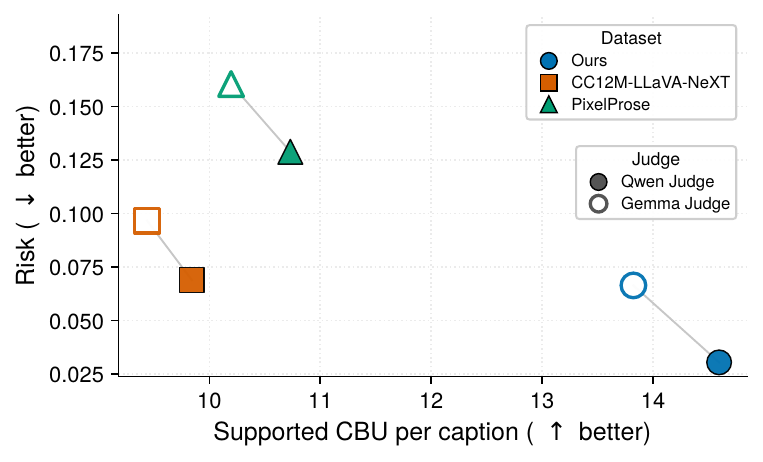}\hfill
\includegraphics[width=0.46\linewidth]{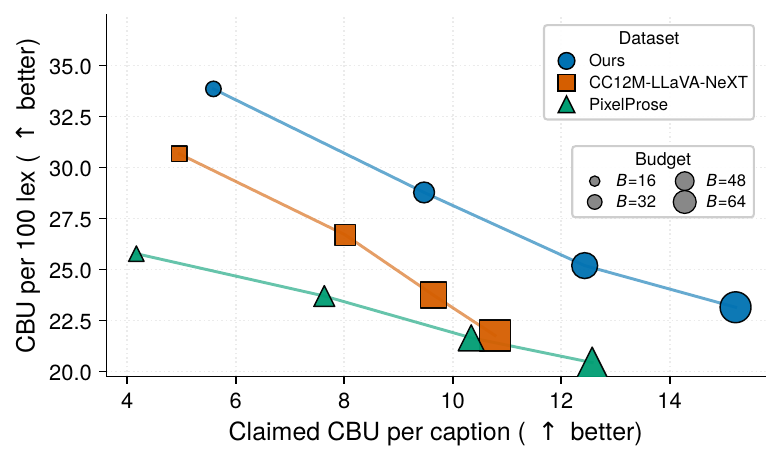}
\caption{\textbf{CC12M long-form surfaces at $B = 64$ and across the budget sweep.} \textit{Left:} supported-yield / risk at $B = 64$; marker shape encodes surface, fill (filled vs hollow) encodes Judge. \textit{Right:} claimed-CBU budget frontier over $B \in \{16,32,48,64\}$ on all $4{,}494$ aligned images; marker size scales with $B$.}
\label{fig:cc12m_frontiers}
\end{figure}

\paragraph{Frontier across three comparisons.}
At $B = 64$ (Figure~\ref{fig:cc12m_frontiers} left, Table~\ref{tab:cc12m_frontier}), Ours holds the highest supported yield and the lowest risk of the long-form surfaces; PixelProse, at $\approx 89$ lex, carries the highest risk ($\rho = 0.129/0.161$ under Qwen/Gemma), and CC12M-LLaVA-NeXT sits between.
The dual-judge gap on supported CBU per caption grows with claim count: $0.12$ on the short Qwen-family reference, $0.40$ on CC12M-LLaVA-NeXT, $0.53$ on PixelProse, and $0.77$ on Ours.

\paragraph{Efficiency and yield read jointly.}
CC12M-Qwen3-VL${}^{\dagger}$ is the short tag-style surface: its $\approx 12$-lex captions reach the highest CBU per 100 lex ($55.84$ against $23.16$ for Ours) and the lowest risk ($\rho = 0.014/0.028$) while stating $6.44$ claims per caption.
One $64$-word window of Ours carries $15.21$ claims, $2.4\times$ a complete CC12M-Qwen3-VL${}^{\dagger}$ caption (Eq.~\ref{eq:cbu_metrics}).

\paragraph{Budget sweep.}
Across $B \in \{16, 32, 48, 64\}$ on all $4{,}494$ aligned images (Figure~\ref{fig:cc12m_frontiers} right), claimed CBU per caption grows and CBU per 100 lex declines on every long-form surface, and Ours leads them on both readings at every budget ($5.59 \to 15.21$ CBU per caption, $33.9 \to 23.2$ per 100 lex).

\begin{table}[t]
\centering
\caption{\textbf{CC12M frontier at $B = 64$, four surfaces under both Judges.} CBU per caption and per $100$ lex come from the Qwen claim extractor $\phi$; each Judge supplies VQA support and risk. The dagger ${}^{\dagger}$ marks the short tag-style surface ($\approx 12$ lex). All columns use the same $4{,}494$ aligned images; per-surface denominators are in Appendix Table~\ref{tab:cc12m_denominators}.}
\label{tab:cc12m_frontier}
\renewcommand{\arraystretch}{1.0}
\setlength{\tabcolsep}{4.0pt}
\scalebox{0.8}{
\begin{tabular}{@{}l rr rr rr@{}}
\toprule
& \multicolumn{2}{c}{\textbf{Surface stats}}
& \multicolumn{2}{c}{\textbf{Qwen Judge}}
& \multicolumn{2}{c}{\textbf{Gemma Judge}} \\
\cmidrule(lr){2-3} \cmidrule(lr){4-5} \cmidrule(l){6-7}
\textbf{Surface}
& \textbf{CBU/cap}\,$\uparrow$
& \textbf{CBU/100lex}
& \textbf{Sup.\ CBU/cap}\,$\uparrow$
& \textbf{Risk}\,$\downarrow$
& \textbf{Sup.\ CBU/cap}\,$\uparrow$
& \textbf{Risk}\,$\downarrow$ \\
\midrule
Ours                              & $\mathbf{15.21}$ & $23.16$ & $\mathbf{14.60}$ & $0.030$ & $\mathbf{13.82}$ & $0.066$ \\
CC12M-LLaVA-NeXT~\citep{conceptualcaptionscc12mllavanext}                  & $10.78$          & $21.77$ & $9.84$           & $0.069$          & $9.44$           & $0.097$ \\
PixelProse~\citep{singla2024pixelprose}                        & $12.57$          & $20.44$ & $10.73$          & $0.129$          & $10.20$          & $0.161$ \\
CC12M-Qwen3-VL${}^{\dagger}$~\citep{cc12mQwen3VL8BHF}      & $6.44$           & $55.84$ & $6.31$           & $0.014$          & $6.19$           & $0.028$ \\
\bottomrule
\end{tabular}}
\end{table}

\subsection{Human Verification Matches the Judges}
\label{sec:human_cbu}

Human annotators, blind to surface identity and machine labels, judged sampled CC12M claims first against the caption window and then against the image (Appendix~\ref{app:human_cbu_details}).
The verification holds $217$ primary judgments on $137$ claims; $80$ claims carry two independent labels, five in each of the $16$ cells of surface group (Ours, pooled references) by claim type.
Table~\ref{tab:human_cbu_descriptive} compares the two image-conditioned
Judges with consensus human labels and reports the human verification result for each
CC12M surface. Qwen and Gemma agree with humans at nearly identical rates.
The annotators find the claims of Ours supported at a higher rate than those of LLaVA-NeXT,
Qwen3-VL-8B, and PixelProse; none of its $111$ judgments is an explicit
\emph{no}, and uncertainty is recorded separately from support.
Human explicit-\emph{no} rates order the long-form surfaces as both Judges' risk does: Ours ($0.0\%$), LLaVA-NeXT ($4.3\%$), PixelProse ($15.4\%$).

\subsection{Auxiliary Embedding-Space Probes}
\label{sec:diagnostics}
The diversity scores Vendi~\citep{friedman2023vendi} and eRank~\citep{roy2007erank}, prompt-to-caption PRDC~\citep{naeem2020prdc}, and caption-to-caption kNN pool captions through a third-party encoder into corpus statistics that follow the encoder.

\paragraph{Within-encoder metric splits.}
Table~\ref{tab:encoder_probes}\,(a) reports DataComp under three encoders.
Vendi and eRank rank Ours and Ref in opposite orders on the same encoder because they weight a shared mean direction differently (Appendix~\ref{app:sensitivity}); on a five-family \texttt{EmbeddingGemma-300M}~\citep{embeddingGemma} grid, Vendi puts Ours lower on CC12M, LAION-pop, and DataComp and higher on PD12M and Danbooru, and eRank reverses the DataComp order.
Caption-to-caption directional kNN cosine is near-symmetric across all three encoders, and prompt-to-caption PRDC gives Ours the higher coverage and density under \texttt{Qwen3-Embedding-4B} and Ref the higher values under \texttt{BGE-M3}, at lower absolute coverage.

\paragraph{Contrastive encoders.}
LongCLIP retrieval on CC12M changes order with caption-side truncation: full mode gives Ours a small R@1 edge with $\approx 32\%$ of its captions over the $248$-token cap, and \texttt{input64} pre-truncation moves CC12M-LLaVA-NeXT to the top in both directions (Table~\ref{tab:encoder_probes}\,(b); Appendix~\ref{app:sensitivity}).
Embedding and retrieval rankings follow the encoder and its truncation policy.

\section{Dataset and Audit-Artifact Release}
\label{sec:release}

The release has two parts: an audited recaption corpus and a reusable audit-artifact bundle. The corpus anchors Sections~\ref{sec:cross}--\ref{sec:cc12m}; the bundle defines the interface for re-running the five-axis profile on future recaptioned corpora, with the released surface as its first audited target.
The corpus is a Hugging Face collection\footnote{\url{https://huggingface.co/collections/BootsofLagrangian/recaptioned-image-text}} of our generated captions for the nine source families of Section~\ref{sec:prelim:targets}, joined by row-level keys to the source releases that hold the images and the reference recaptions.
The code repository carries the audit pipeline, the CC12M and DataComp naive-control surfaces of Section~\ref{sec:captioner_control}, and the result summaries behind every table (Appendix~\ref{app:release_manifest}); caption generation cost is $\approx 12.7$k H200 GPU-hours (Appendix~\ref{app:compute}).
Captions are released as generated and caption-only, so each training pipeline applies its own personal-information, safety, NSFW, and dedup filters.
\paragraph{Auditing other recaptioned surfaces.}
The bundle is corpus-agnostic: a third party supplies caption text, source-row identifiers, and (for the image-conditioned axis) paired image access; the released tokenizer, prompt pools, CBU schema, VQA template, label schema, and manifest are then reused. Text-side axes need captions alone. Record schema, reproduction workflow, and Croissant 1.1 / Responsible-AI coverage are in Appendix~\ref{app:release_manifest}.

\section{Limitations}
\label{sec:limitations}
The audit measures caption--image supervision before any T2I training pass.
Image-output quality additionally depends on the text encoder, on caption length sampling, dropout, and persona augmentation, and on the generator and its training budget; $99.89$--$99.99\%$ of Ours captions exceed the $77$-token CLIP context (Appendix Table~\ref{tab:encoder_truncation}), so an image-output comparison measures the encoder and augmentation policy together with the corpus.
Image-side safety, NSFW, and dedup filtering belong to the training pipeline (Section~\ref{sec:release}).
Risk is the share of claimed CBUs a Judge labels unsupported and aggregates captioner, extractor, and Judge errors; the Gemma Judge varies the Judge family while the extractor $\phi$ stays on the Qwen checkpoint.
Human judgments anchor the image-conditioned axis on sampled CC12M claims (Section~\ref{sec:human_cbu}), and the two Judges agree least on count and text-rendering claims ($85.1\%$ and $87.6\%$, Appendix Table~\ref{tab:vqa_by_type}).
The axis scores the claims a caption states; an image-first inventory measures the content a caption omits.
The paired audit spans five source families, the four release-only families extend the released corpus, and the prompt pools cover natural-language T2I prompts.
Additional source families, booru-style prompt pools, and larger human-anchored samples are the next targets of the same interface.

\begin{table}[t!]
\centering
\caption{\textbf{Human verification of image support.} Panel (a): design-weighted nominal exact agreement on $43$ resolved claims, estimate
$\pm$ standard deviation over $10{,}000$ image-cluster bootstrap resamples. Panel (b): unweighted
distribution of all $217$ primary image judgments; uncertainty
remains separate from support, and ``other'' is \emph{not visual} or
\emph{prefer not to answer}.}
\label{tab:human_cbu_descriptive}
\footnotesize
\renewcommand{\arraystretch}{1.0}
\setlength{\tabcolsep}{5.0pt}
\begin{tabular}{@{}lrrrrr@{}}
\toprule
\multicolumn{6}{@{}l}{\textbf{(a) Judge--human agreement}} \\
\addlinespace[0.15em]
\textbf{Judge}
& \textbf{$n$}
& \multicolumn{2}{r}{\textbf{Overall}}
& \textbf{Ours}
& \textbf{Pooled refs.} \\
\midrule
Qwen  & $43$ & \multicolumn{2}{r}{$84.8 \pm 9.7\%$} & $87.9 \pm 10.9\%$ & $82.5 \pm 14.6\%$ \\
Gemma & $43$ & \multicolumn{2}{r}{$84.2 \pm 9.7\%$} & $89.0 \pm 10.8\%$ & $80.7 \pm 14.7\%$ \\
\midrule
\multicolumn{6}{@{}l}{\textbf{(b) Human image support}} \\
\addlinespace[0.15em]
\textbf{Surface}
& \textbf{$n$}
& \textbf{Yes}
& \textbf{Uncertain}
& \textbf{Explicit no}
& \textbf{Other} \\
\midrule
\textbf{Ours}
& $111$ & $\mathbf{87\;(78.4\%)}$ & $22\;(19.8\%)$
& $\mathbf{0\;(0.0\%)}$ & $2\;(1.8\%)$ \\
LLaVA-NeXT
& $47$ & $33\;(70.2\%)$ & $10\;(21.3\%)$
& $2\;(4.3\%)$ & $2\;(4.3\%)$ \\
Qwen3-VL-8B
& $20$ & $14\;(70.0\%)$ & $5\;(25.0\%)$
& $1\;(5.0\%)$ & $0\;(0.0\%)$ \\
PixelProse
& $39$ & $23\;(59.0\%)$ & $9\;(23.1\%)$
& $6\;(15.4\%)$ & $1\;(2.6\%)$ \\
\bottomrule
\end{tabular}
\end{table}

\begin{table}[t!]
\centering
\caption{\textbf{Encoder probes follow the encoder and its truncation.} (a) DataComp, Ours and Ref under three text encoders; Coverage is prompt-to-caption PRDC. (b) CC12M LongCLIP R@1 with full captions and with captions cut to $B = 64$ (\texttt{input64}). Full tables in Appendix~\ref{app:sensitivity}.}
\label{tab:encoder_probes}
\footnotesize
\renewcommand{\arraystretch}{1.0}
\setlength{\tabcolsep}{2.4pt}
\begin{minipage}[t]{0.53\linewidth}
\centering
\begin{tabular}{@{}l rr rr rr@{}}
\toprule
\multicolumn{7}{@{}l}{\textbf{(a) DataComp text-space probes}} \\
\addlinespace[0.15em]
& \multicolumn{2}{c}{\textbf{Vendi}\,$\uparrow$} & \multicolumn{2}{c}{\textbf{eRank}\,$\uparrow$} & \multicolumn{2}{c}{\textbf{Coverage}\,$\uparrow$} \\
\cmidrule(lr){2-3} \cmidrule(lr){4-5} \cmidrule(l){6-7}
\textbf{Encoder} & Ours & Ref & Ours & Ref & Ours & Ref \\
\midrule
Qwen3-Emb-4B & $61.40$ & $\mathbf{62.57}$ & $\mathbf{337.2}$ & $319.5$ & $\mathbf{0.743}$ & $0.357$ \\
Qwen3-Emb-8B & $76.33$ & $\mathbf{89.25}$ & $\mathbf{424.0}$ & $378.5$ & -- & -- \\
BGE-M3       & $25.39$ & $\mathbf{34.21}$ & $\mathbf{255.1}$ & $235.5$ & $0.017$ & $\mathbf{0.050}$ \\
\bottomrule
\end{tabular}
\end{minipage}\hfill
\begin{minipage}[t]{0.43\linewidth}
\centering
\begin{tabular}{@{}l rr @{\hspace{1.0em}} rr@{}}
\toprule
\multicolumn{5}{@{}l}{\textbf{(b) CC12M LongCLIP R@1}} \\
\addlinespace[0.15em]
& \multicolumn{2}{c}{\textbf{I2T}\,$\uparrow$} & \multicolumn{2}{c}{\textbf{T2I}\,$\uparrow$} \\
\cmidrule(lr){2-3} \cmidrule(l){4-5}
\textbf{Surface} & full & input64 & full & input64 \\
\midrule
Ours       & $\mathbf{0.908}$ & $0.859$ & $\mathbf{0.902}$ & $0.850$ \\
LLaVA-NeXT & $0.907$ & $\mathbf{0.897}$ & $0.899$ & $\mathbf{0.888}$ \\
PixelProse & $0.890$ & $0.869$ & $0.874$ & $0.856$ \\
\bottomrule
\end{tabular}
\end{minipage}
\end{table}

\section{Conclusion}
\label{sec:conclusion}
A recaptioned image-text corpus is a supervision distribution set by a captioning policy, a captioner, and a source corpus.
We introduced a reusable matched-budget audit framework that reports a five-axis profile over prompt-side coverage, image-conditioned faithfulness, and caption-surface health at a fixed text budget, with controllable basic units as the common claim set behind claimed density, supported yield, and unsupported risk.
Instantiating the same framework on seven paired comparisons, the released surface is the high-yield, lower-risk side of every cross-corpus pair under both Judges; CC12M traces a long-vs-dense frontier preserved across two Judges and four budgets; and a same-captioner policy ablation locates the cross-corpus gap on the policy axis.
The released corpus and audit-artifact bundle let third parties reproduce the reported profile and audit future recaptioned corpora under the same measurement interface.

\begin{ack}
This work was partly supported by the Institute of Information \& Communications Technology Planning \& Evaluation (IITP) grant funded by the Korean government (MSIT) (No. RS-2021-II211343, Artificial Intelligence Graduate School Program (Seoul National University), No. RS-2026-25522885, Development of a World Foundation Model for Training and Development of Physical AI Systems, No. RS-2026-25512061, Development of Large Action Model-Based Autonomous Digital Twin Operation Technology for Proactive Problem Solving), the National Research Foundation of Korea (NRF) grant funded by the Korean government (MSIT) (No. RS-2024-00354218), and the Technology Innovation Program (RS-2025-25456760, Development of a humanoid robot specialized in chemical processes based on AI foundation model) funded by the Ministry of Trade, Industry and Resources (MOTIR, Korea).\end{ack}

\clearpage
\appendix
\section{Notation and Audit Vocabulary}
\label{app:glossary}

Table~\ref{tab:glossary} lists the symbols and terms of the audit, grouped by the part of the framework that uses them.

\begin{table}[h]
\centering
\caption{\textbf{Notation and audit vocabulary.}}
\label{tab:glossary}
\small
\renewcommand{\arraystretch}{1.12}
\begin{tabularx}{\linewidth}{@{}l >{\raggedright\arraybackslash}X@{}}
\toprule
\textbf{Symbol or term} & \textbf{Meaning} \\
\midrule
\multicolumn{2}{@{}l}{\emph{Supervision distribution}} \\
$C$, $x$ & source image corpus and one of its images \\
$V_c$, $\pi$ & captioner and captioning policy \\
$D_{\pi,V_c,C}$ & caption--image pairs written by $V_c$ under $\pi$ on $C$; the audit target \\
$D_c$, $D_{cx}$ & caption marginal and caption--image joint \\
$D^{\star}$ (Ours) & the released corpus, the first audit target \\
Surface & one caption set over the same source rows \\
Describe-style frame & caption-only scaffolding such as \texttt{The image shows\dots} \\
\addlinespace
\multicolumn{2}{@{}l}{\emph{Budget}} \\
$B$ & text budget ($B = 64$): lexical units for text statistics, words for claim extraction \\
Lexical unit (lex) & regex word unit after Unicode normalization \\
$c_{\le B}$, $\ell(\cdot)$ & first $B$ whitespace-delimited words of caption $c$ and their length in lexical units \\
$B$-eligibility & share of captions reaching $B$ lexical units \\
\addlinespace
\multicolumn{2}{@{}l}{\emph{Claims}} \\
CBU & controllable basic unit, one atomic visual claim of one of eight types \\
$\phi$ & claim extractor, reading the caption alone \\
$V_J$ (Judge) & image-conditioned VLM that re-asks each CBU: Qwen or Gemma \\
Claimed CBU/cap & CBUs that $\phi$ finds in the window, per caption \\
CBU/100\,lex & claimed CBUs per $100$ lexical units of the window (Eq.~\ref{eq:cbu_metrics}) \\
Supported CBU/cap & CBUs the Judge answers \emph{yes} for, per caption \\
Risk $\rho$ & share of claimed CBUs the Judge answers \emph{no} for \\
\addlinespace
\multicolumn{2}{@{}l}{\emph{Prompt side}} \\
$p$, $P$ & user prompt and prompt distribution \\
Prompt pool & public prompt set standing in for user prompts \\
Prompt-mass support & share of pool prompt mass whose $n$-grams appear in the captions \\
$n$-gram JSD & Jensen--Shannon divergence between caption and pool $n$-grams \\
\addlinespace
\multicolumn{2}{@{}l}{\emph{Surface health}} \\
Opener rate & share of captions whose opening matches the describe-style regex catalog \\
Top-100 prefix mass & share of captions whose first five lexical units are among the $100$ most frequent prefixes \\
Content prefix mass & the same share after leading articles and frame words are dropped \\
Distinct-3 & unique $3$-grams over all $3$-grams of the corpus \\
\bottomrule
\end{tabularx}
\end{table}

\section{Computational Resources}
\label{app:compute}

The H200 GPU-hour budget covers two stages: VLM caption generation for the released corpus and sampled VLM evaluation for the matched-budget profile. Caption generation dominates; sampled audit and encoder-probe costs add a small additional contribution.

\paragraph{Caption generation.}
Production recaptioning runs eight H200 GPUs as data-parallel vLLM workers using \texttt{Qwen/Qwen3.5-35B-A3B-FP8} (MoE 35B/A3B active). At the measured steady-state throughput of $\approx 90$ accepted captions per second on $8 \times \text{H200}$, generating the $514.6$M caption rows behind the $\approx 490$M image identities of the release (identities counted within each family; rows repeated by resumed runs and alternative caption variants included) corresponds to approximately $\mathbf{12.7\text{k H200 GPU-hours}}$ ($514.6\mathrm{M} / 90\,\mathrm{rows\,s^{-1}} \times 8\,\mathrm{GPUs}\,/\,3600$). A throughput sensitivity band of $80$--$115$ accepted captions per second gives a range of approximately $9.9$--$14.3$k H200 GPU-hours.

\paragraph{Sampled VLM audits.}
Image-conditioned audits use \texttt{Qwen/Qwen3.5-397B-A17B-FP8} (MoE 397B/A17B active) on tensor-parallel H200 pods ($2$ or $4$ GPUs); the \texttt{Gemma-4-31B-IT} judge runs at a comparable order of cost on its own pod. Table~\ref{tab:compute_audit} lists the three audit stages with their request and compute scales.

\begin{table}[h]
\centering
\small
\caption{Sampled VLM audit stages, request scales, and approximate compute orders. The two image-conditioned stages dominate; text-only extraction is one to two orders smaller.}
\begin{tabular}{@{}l l r r@{}}
\toprule
\textbf{Audit stage} & \textbf{Input} & \textbf{Requests} & \textbf{H200 GPU-hours} \\
\midrule
ClaimedCBU@64  & text           & $\approx 100$k & $\approx 1$ \\
GroundedCBU@64 & text + image   & $\approx 70$k  & tens \\
CBU-VQA        & question + image & $\approx 100$k & tens \\
\bottomrule
\end{tabular}
\label{tab:compute_audit}
\end{table}

Embedding-encoder probes (PRDC, kNN, Vendi, eRank) and LongCLIP retrieval add a smaller contribution on the order of low tens of H200 GPU-hours combined.

\paragraph{Total.}
The aggregate compute cost is approximately $\mathbf{12.7\text{k H200 GPU-hours}}$, dominated by caption generation; the sampled audit and encoder probes together contribute on the order of $10^2$ H200 GPU-hours. The result summaries in the code repository record the request counts of each stage.

\section{Protocol Details}
\label{app:protocol}

\paragraph{CBU schema and VLM request templates.}
Claim extraction uses the CBU type taxonomy and image verification the supported / unsupported / uncertain reply schema (Section~\ref{sec:protocol:axes}); Appendix~\ref{app:vlm_templates} reproduces the request templates verbatim: photorealistic and anime caption generation, the captioner-control naive prompt, the claim extractor $\phi$, and the image-conditioned Judge.

\paragraph{Prompt-pool definitions.}
Table~\ref{tab:prompt_pools} lists the seven prompt-reference pools used per slice in the prompt-pool support axis: four FLUX-style community prompt sets, one Stable Diffusion prompt set, one SDXL-refiner set, and Pick-a-Pic preference rankings. We omit DiffusionDB-derived deduplications to avoid double-counting upstream content; the released audit-artifact bundle ships the eight-pool variant for sensitivity comparison.
Each pool contributes up to $1$M prompts and is archived under a stable name; each slice contributes $250{,}000$ caption pairs ($114{,}621$ on CC12M--Qwen3-VL).

\begin{table}[h]
\centering
\small
\caption{The seven prompt-reference pools per slice behind the prompt-pool support axis (Section~\ref{sec:protocol:axes}).
Each pool is a public Hugging Face community prompt set for a named T2I model family or the cross-model Pick-a-Pic preference rankings, and results are reported per pool.}
\label{tab:prompt_pools}
\renewcommand{\arraystretch}{1.10}
\begin{tabularx}{\linewidth}{@{}l l X@{}}
\toprule
\textbf{Pool name} & \textbf{Target} & \textbf{Source / hosting} \\
\midrule
\texttt{civitai\_flux\_prompts\_aconexx}  & FLUX             & {\scriptsize\url{https://huggingface.co/datasets/Aconexx/CivitAI-Flux-Prompts}} \\
\texttt{flux\_improved\_k\_mktr}          & FLUX             & {\scriptsize\url{https://huggingface.co/datasets/k-mktr/improved-flux-prompts}} \\
\texttt{flux\_prompts\_chrisgoringe}      & FLUX             & {\scriptsize\url{https://huggingface.co/datasets/ChrisGoringe/flux_prompts}} \\
\texttt{flux\_prompts\_regpeter}          & FLUX             & {\scriptsize\url{https://huggingface.co/datasets/regpeter/flux_prompts}} \\
\texttt{sd\_prompts\_2m\_andyyang}        & Stable Diffusion & {\scriptsize\url{https://huggingface.co/datasets/andyyang/stable_diffusion_prompts_2m}} \\
\texttt{sdxl\_refiner\_prompts\_falah}    & SDXL refiner     & {\scriptsize\url{https://huggingface.co/datasets/Falah/1M_SDXL_Refiner_Prompts}} \\
\texttt{pickapic\_rankings}               & cross-model      & Pick-a-Pic preference rankings~\citep{kirstain2023pickapic} \\
\bottomrule
\end{tabularx}
\end{table}

\begin{figure}[t]
\centering
\includegraphics[width=\linewidth]{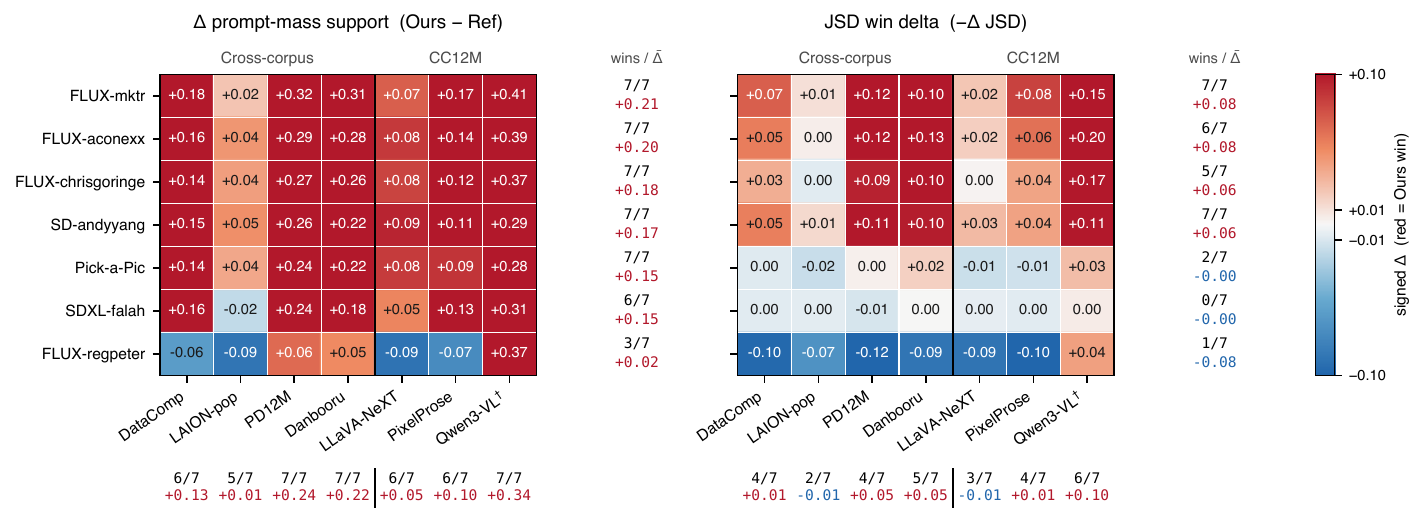}
\caption{\textbf{Per-pool prompt-support direction across paired comparisons.}
Each cell is the mean Ours-minus-Ref delta over $250{,}000$ caption pairs per comparison ($114{,}621$ on CC12M--Qwen3-VL) at $B = 64$ lexical units (Section~\ref{sec:protocol:desiderata}); rows share a sort by Ours-minus-Ref prompt-mass mean.
\emph{Left:} $\Delta$ prompt-mass support (Ours $-$ Ref); positive means Ours covers more pool $n$-gram mass.
\emph{Right:} direction-aligned JSD, plotted as $-\Delta\mathrm{JSD} = -(\mathrm{JSD}_{\mathrm{Ours}} - \mathrm{JSD}_{\mathrm{Ref}})$, so positive means Ours is closer to the pool $n$-gram distribution, and red marks the favorable direction in both panels.
Color scale is symmetric on $[-0.10, +0.10]$ through a near-neutral zero band; measured deltas with $|\Delta| < 0.01$ print as unsigned $0.00$, and saturated cells with $|\Delta| > 0.10$ print the full signed value in bold.
Right margin per panel: per-pool wins / $7$ paired comparisons, plus the mean delta $\bar\Delta$ across those comparisons.
Bottom margin per panel: per-slice wins / $7$ pools, plus mean $\bar\Delta$.
The vertical separator splits four cross-corpus paired comparisons (Section~\ref{sec:cross}) from the three CC12M case-study references (Section~\ref{sec:cc12m}).}
\label{fig:per_pool_heatmap}
\end{figure}

\paragraph{Lexical tokenization and budget.}
Text statistics read the first $64$ regex-tokenized lexical units (\texttt{[\^{}\textbackslash W\_]+(?:'[\^{}\textbackslash W\_]+)*}) after Unicode normalization.
The claim extractor receives the first $B$ whitespace-delimited words of each caption, and the CBU/100\,lex denominator counts regex lexical units inside that window; the two counts differ where inner punctuation such as a hyphen splits a whitespace word into several lexical units or a standalone symbol occupies a word slot. Cutting the window at $64$ lexical units instead preserves the order of the four CC12M surfaces, changes claimed CBU per caption by at most $3.1\%$ (Ours $15.21 \to 14.75$), and changes CBU/100\,lex by less than $0.15$.
Encoder-specific tokenizer behavior (CLIP 77-token cap, T5 512-token cap, EmbeddingGemma) is documented separately as encoder sensitivity; the encoder-token truncation diagnostic is reported in Appendix Table~\ref{tab:encoder_truncation}.

\paragraph{Artifact column-name compatibility.}
Artifact CSV/TSV files keep the column names \texttt{avg\_tokens}, \texttt{avg\_lexical\_tokens}, \texttt{cov64}, \texttt{elig64}, and \texttt{cbu\_100tok} for join compatibility; lengths, eligibility, and claim-density denominators in them count regex lexical units, independent of BPE, SentencePiece, and CLIP tokenizers.
The lexical tokenizer \texttt{[\^{}\textbackslash W\_]+(?:'[\^{}\textbackslash W\_]+)*} counts word-like units and leaves punctuation uncounted.
The column \texttt{cbu\_100tok} holds $\mathrm{CBU/100lex} = 100 \times \sum (\text{deduplicated claimed CBU}) / \sum(\text{retained lexical units in evaluated captions})$, where a caption shorter than $B$ contributes its actual retained length.

\section{Per-Slice Audit Results}
\label{app:numbers}

\paragraph{VQA question denominators.}
Table~\ref{tab:vqa_questions} lists the per-surface VQA response and question counts behind every Section~\ref{sec:cross} VQA cell.
\emph{Responses} is the number of captions audited (one VLM response per audited caption); \emph{Questions} is the total number of CBU questions issued across those captions (sum over captions of CBU count at $B = 64$).
The Risk column follows the definition of Section~\ref{sec:protocol:desiderata} at $B = 64$.

\begin{table}[h]
\centering
\caption{\textbf{VQA question denominators and $B$-eligibility behind Section~\ref{sec:cross}.} Qwen Judge at $B = 64$, on the requests answered by both Judges. $B$-elig = share of captions reaching $B$ lexical units; the short captions of PD12M-released, Recap-DataComp, and Danbooru-Florence fill part of the window. Q/cap = average CBU per audited caption.}
\label{tab:vqa_questions}
\scalebox{0.9}{
\begin{tabularx}{\linewidth}{@{}l X r r r r r@{}}
\toprule
\textbf{Dataset} & \textbf{Surface} & \textbf{$B$-elig$\uparrow$} & \textbf{Responses} & \textbf{Questions} & \textbf{Q/cap} & \textbf{Risk} \\
\midrule
DataComp  & Ours & $1.00$ & $4{,}771$ & $68{,}933$ & $14.45$ & $0.035$ \\
DataComp  & Recap-DataComp \citep{li2024recapdatacomp} & $0.26$ & $4{,}914$ & $51{,}356$ & $10.45$ & $0.177$ \\
Danbooru  & Ours & $1.00$ & $4{,}991$ & $71{,}491$ & $14.32$ & $0.058$ \\
Danbooru  & Danbooru-Florence \citep{xiao2024florence2,danbooruFlorence2HF} & $0.17$ & $4{,}969$ & $40{,}755$ & $8.20$ & $0.217$ \\
LAION-pop & Ours & $1.00$ & $4{,}962$ & $73{,}489$ & $14.81$ & $0.031$ \\
LAION-pop & LAION-pop-Llama \citep{laionPopLlama32HF} & $1.00$ & $4{,}946$ & $58{,}903$ & $11.91$ & $0.077$ \\
PD12M     & Ours & $1.00$ & $4{,}955$ & $74{,}392$ & $15.01$ & $0.034$ \\
PD12M     & PD12M released \citep{pd12mFullHF,meyer2024pd12m} & $0.07$ & $4{,}989$ & $48{,}825$ & $9.79$ & $0.103$ \\
\bottomrule
\end{tabularx}}
\end{table}

\paragraph{CC12M denominators.}
Table~\ref{tab:cc12m_denominators} traces every CC12M surface from the $4{,}494$ images that all four surfaces share to the VQA cells of Tables~\ref{tab:captioner_control_cbu} and~\ref{tab:cc12m_frontier}.
Under schema-constrained decoding, $99$ of the $17{,}976$ extraction requests ($0.55\%$) emit whitespace until the $4{,}096$-token output cap and return no parsable record; claimed CBU/cap and CBU/100\,lex average over the valid extractions.
Captions with zero claims issue no VQA question, so each surface's VQA responses equal its valid extractions minus its zero-claim captions, and both Judges answer the identical question set.
The budget sweep of Figure~\ref{fig:cc12m_frontiers} (right) takes $B = 64$ from the same extraction and $B \in \{16, 32, 48\}$ from two extraction runs with the same checkpoint, prompts, and greedy schema-constrained decoding, one over the first $1{,}000$ images and one over the remaining $3{,}494$; on $800$ requests repeated in both runs, claimed CBU per caption agrees within $1.8\%$ for every surface.

\begin{table}[h]
\centering
\caption{\textbf{CC12M denominators at $B = 64$.} Aligned: images shared by all four surfaces. Valid: parsable extractions. Cap: extractions that reached the $4{,}096$-token output cap. Zero: valid extractions with no claim. VQA: audited captions; Q: CBU questions, identical for both Judges.}
\label{tab:cc12m_denominators}
\footnotesize
\renewcommand{\arraystretch}{1.08}
\setlength{\tabcolsep}{3.5pt}
\begin{tabular}{@{}l rrrr rr rr@{}}
\toprule
& & & & & & & \multicolumn{2}{c}{\textbf{Risk}} \\
\cmidrule(l){8-9}
\textbf{Surface} & \textbf{Aligned} & \textbf{Valid} & \textbf{Cap} & \textbf{Zero} & \textbf{VQA} & \textbf{Q} & \textbf{Qwen} & \textbf{Gemma} \\
\midrule
Ours & $4{,}494$ & $4{,}470$ & $24$ & $3$ & $4{,}467$ & $68{,}019$ & $0.030$ & $0.066$ \\
CC12M-LLaVA-NeXT & $4{,}494$ & $4{,}459$ & $35$ & $6$ & $4{,}453$ & $48{,}092$ & $0.069$ & $0.097$ \\
PixelProse & $4{,}494$ & $4{,}457$ & $37$ & $8$ & $4{,}449$ & $56{,}043$ & $0.129$ & $0.161$ \\
CC12M-Qwen3-VL${}^{\dagger}$ & $4{,}494$ & $4{,}491$ & $3$ & $1$ & $4{,}490$ & $28{,}911$ & $0.014$ & $0.028$ \\
\bottomrule
\end{tabular}
\end{table}

\paragraph{Surface-descriptor measurement.}
\label{app:phenomenon_method}
The four full-caption descriptors of Section~\ref{sec:prelim:gap} are computed on the paired slice of each surface, capped at $1$M captions ($42{,}231$ on LAION-pop, $114{,}621$ on CC12M--Qwen3-VL, $729{,}237$ on PixelProse, $960{,}394$ on CC12M-LLaVA-NeXT, and $1$M on DataComp, PD12M, and Danbooru).
\emph{Opener rate} is the fraction of captions matching a regex catalog of third-person caption-style frames; the full catalog ships with the released audit-artifact bundle, and Section~\ref{sec:prelim:gap} shows its principal form.
\emph{Raw-prefix top-100 mass} takes the first five lexical units of each caption as its prefix and reports the share of captions whose prefix is among the $100$ most frequent prefixes of the surface.
\emph{Content-prefix top-100 mass} repeats the count after dropping leading articles, demonstratives, and frame words (\texttt{image}, \texttt{photo}, \texttt{shows}, \texttt{depicts}, \dots) from each caption, isolating concentration that survives boilerplate removal.
\emph{Distinct-3-gram rate} is the fraction of unique 3-grams over all 3-grams in the corpus after lowercasing and Unicode normalization; lower is more repetitive.

\paragraph{Surface descriptors by source family.}
Table~\ref{tab:phenomenon_full} expands Table~\ref{tab:phenomenon_meta_opener} with per-pair Ours and Reference rows for the four surface descriptors.

\begin{table}[h]
\centering
\caption{Surface descriptors by source family (companion to Table~\ref{tab:phenomenon_meta_opener}). Ref = the comparison-surface release named in the Pair column.}
\label{tab:phenomenon_full}
\renewcommand{\arraystretch}{1.10}
\setlength{\tabcolsep}{4pt}
\scalebox{0.8}{
\begin{tabular}{l l l r r r r}
\toprule
\textbf{Source} & \textbf{Pair} & \textbf{Surface} & \textbf{Opener} & \textbf{Top-100 raw} & \textbf{Top-100 content} & \textbf{Distinct-3} \\
\midrule
DataComp   & Recap-DataComp           & Ref  & $0.36\%$   & $27.99\%$ & $17.53\%$ & $11.49\%$ \\
           &                           & Ours & $2.00\%$   & $7.69\%$  & $5.31\%$  & $19.26\%$ \\
\addlinespace
CC12M      & CC12M-LLaVA-NeXT          & Ref  & $0.24\%$   & $22.97\%$ & $12.39\%$ & $13.81\%$ \\
           &                           & Ours & $1.87\%$   & $10.57\%$ & $6.83\%$  & $18.05\%$ \\
           & PixelProse                & Ref  & $100.00\%$ & $65.07\%$ & $8.25\%$  & $12.56\%$ \\
           &                           & Ours & $1.86\%$   & $10.51\%$ & $6.79\%$  & $19.54\%$ \\
           & CC12M-Qwen3-VL${}^{\dagger}$ & Ref  & $0.00\%$ & $9.40\%$  & $5.87\%$  & $37.04\%$ \\
           &                           & Ours & $1.39\%$   & $11.61\%$ & $7.15\%$  & $28.28\%$ \\
\addlinespace
LAION-pop  & LAION-pop-Llama           & Ref  & $0.78\%$   & $23.68\%$ & $14.54\%$ & $24.83\%$ \\
           &                           & Ours & $1.27\%$   & $15.82\%$ & $10.34\%$ & $36.87\%$ \\
\addlinespace
PD12M      & PD12M released            & Ref  & $99.98\%$  & $90.73\%$ & $43.10\%$ & $2.40\%$  \\
           &                           & Ours & $2.28\%$   & $10.18\%$ & $6.65\%$  & $15.78\%$ \\
\addlinespace
Danbooru   & Danbooru-Florence         & Ref  & $77.29\%$  & $99.46\%$ & $61.10\%$ & $1.34\%$  \\
           &                           & Ours & $2.03\%$   & $4.30\%$  & $2.36\%$  & $13.49\%$ \\
\bottomrule
\end{tabular}}
\end{table}

\paragraph{CBU category breakdowns.}
Per-category CBU counts (object, attribute, relation, count, style, camera, lighting, text-rendering) per surface and per slice are released with the audit-artifact bundle (Appendix~\ref{app:release_manifest}).
Attribute is the most consistent driver of CBU growth on the released surface; style and text-rendering decrease relative to several comparisons.
On PD12M, Ours states slightly fewer object CBUs than the released reference.

\paragraph{Claim-type robustness.}
Excluding count and relation claims, Ours raises supported CBU per caption by $+2.66$ to $+5.22$ and lowers risk in all eight (pair, judge) cells (Table~\ref{tab:excl_count_relation}).
Table~\ref{tab:vqa_by_type} breaks support and risk down by claim type: count and text-rendering carry the highest risk under both Judges and the lowest Qwen--Gemma exact agreement on CC12M, while Ours has lower risk than the pooled references in every type under both Judges.

\begin{table}[h]
\centering
\caption{\textbf{Cross-corpus results on six claim types ($B = 64$).} The six-type subset leaves out count and relation. Cells $a \to b$ report Ref $\to$ Ours (bold = Ours); $\Delta$ is the supported-yield gain on the subset.}
\label{tab:excl_count_relation}
\footnotesize
\renewcommand{\arraystretch}{1.08}
\setlength{\tabcolsep}{4.0pt}
\begin{tabular}{@{}ll cc r c@{}}
\toprule
& & \multicolumn{2}{c}{\textbf{Sup.\ CBU/cap}$\,\uparrow$} & & \\
\cmidrule(lr){3-4}
\textbf{Dataset} & \textbf{Judge} & \textbf{all types} & \textbf{six types} & $\boldsymbol{\Delta}$ & \textbf{Risk, six types}$\,\downarrow$ \\
\midrule
DataComp & Qwen & $8.49 \to \mathbf{13.73}$ & $6.83 \to \mathbf{11.05}$ & $+4.22$ & $0.165 \to \mathbf{0.032}$ \\
 & Gemma & $8.00 \to \mathbf{12.94}$ & $6.47 \to \mathbf{10.46}$ & $+3.99$ & $0.203 \to \mathbf{0.073}$ \\
LAION-pop & Qwen & $10.80 \to \mathbf{14.22}$ & $8.87 \to \mathbf{11.57}$ & $+2.70$ & $0.064 \to \mathbf{0.027}$ \\
 & Gemma & $10.22 \to \mathbf{13.61}$ & $8.43 \to \mathbf{11.10}$ & $+2.66$ & $0.095 \to \mathbf{0.053}$ \\
PD12M & Qwen & $8.61 \to \mathbf{14.29}$ & $6.37 \to \mathbf{11.36}$ & $+4.98$ & $0.089 \to \mathbf{0.029}$ \\
 & Gemma & $8.23 \to \mathbf{13.53}$ & $6.11 \to \mathbf{10.76}$ & $+4.65$ & $0.112 \to \mathbf{0.058}$ \\
Danbooru & Qwen & $6.38 \to \mathbf{12.74}$ & $4.99 \to \mathbf{10.20}$ & $+5.22$ & $0.221 \to \mathbf{0.057}$ \\
 & Gemma & $6.15 \to \mathbf{11.85}$ & $4.88 \to \mathbf{9.48}$ & $+4.60$ & $0.224 \to \mathbf{0.091}$ \\
\bottomrule
\end{tabular}
\end{table}

\begin{table}[h]
\centering
\caption{\textbf{Image support and risk by claim type ($B = 64$).} Rates are shares of CBU questions answered \emph{yes} (Sup.) or \emph{no} (Risk), pooled over the four cross-corpus pairs and the CC12M comparison; Refs pools every reference surface. Agree.\ is exact Qwen--Gemma answer agreement on the identical CC12M question set.}
\label{tab:vqa_by_type}
\footnotesize
\renewcommand{\arraystretch}{1.08}
\setlength{\tabcolsep}{3.5pt}
\begin{tabular}{@{}l cc cc cc cc r@{}}
\toprule
& \multicolumn{4}{c}{\textbf{Qwen Judge}} & \multicolumn{4}{c}{\textbf{Gemma Judge}} & \\
\cmidrule(lr){2-5} \cmidrule(lr){6-9}
& \multicolumn{2}{c}{\textbf{Ours}} & \multicolumn{2}{c}{\textbf{Refs}} & \multicolumn{2}{c}{\textbf{Ours}} & \multicolumn{2}{c}{\textbf{Refs}} & \\
\cmidrule(lr){2-3} \cmidrule(lr){4-5} \cmidrule(lr){6-7} \cmidrule(lr){8-9}
\textbf{Type} & Sup. & Risk & Sup. & Risk & Sup. & Risk & Sup. & Risk & \textbf{Agree.} \\
\midrule
object & $0.931$ & $0.029$ & $0.914$ & $0.072$ & $0.888$ & $0.053$ & $0.881$ & $0.098$ & $94.4\%$ \\
attribute & $0.947$ & $0.036$ & $0.849$ & $0.131$ & $0.890$ & $0.073$ & $0.801$ & $0.161$ & $91.3\%$ \\
relation & $0.952$ & $0.042$ & $0.855$ & $0.137$ & $0.898$ & $0.091$ & $0.802$ & $0.186$ & $91.2\%$ \\
count & $0.889$ & $0.106$ & $0.734$ & $0.251$ & $0.809$ & $0.178$ & $0.679$ & $0.299$ & $85.1\%$ \\
style & $0.953$ & $0.038$ & $0.895$ & $0.092$ & $0.941$ & $0.044$ & $0.933$ & $0.045$ & $96.5\%$ \\
camera & $0.956$ & $0.043$ & $0.897$ & $0.099$ & $0.930$ & $0.069$ & $0.873$ & $0.120$ & $93.8\%$ \\
lighting & $0.951$ & $0.037$ & $0.905$ & $0.085$ & $0.938$ & $0.047$ & $0.896$ & $0.089$ & $94.8\%$ \\
text-rendering & $0.931$ & $0.063$ & $0.690$ & $0.305$ & $0.814$ & $0.172$ & $0.639$ & $0.352$ & $87.6\%$ \\
\midrule
all types & \multicolumn{8}{c}{} & $92.4\%$ \\
\bottomrule
\end{tabular}
\end{table}

\paragraph{Bootstrap standard deviations.}
Table~\ref{tab:bootstrap_ci} reports every VQA cell of Sections~\ref{sec:cross}--\ref{sec:cc12m} as mean $\pm$ standard deviation over a caption-level bootstrap ($2{,}000$ resamples).

\begin{table}[h]
\centering
\caption{\textbf{VQA cells at $B = 64$ as mean $\pm$ bootstrap standard deviation ($2{,}000$ caption-level resamples).} Sup CBU/cap and Risk per surface under the Qwen Judge and the Gemma Judge; Risk is unsupported / claimed CBU. The control rows belong to Table~\ref{tab:captioner_control_cbu}; Naive (greedy) decodes the naive prompt at temperature $0$.}
\label{tab:bootstrap_ci}
\footnotesize
\renewcommand{\arraystretch}{1.10}
\setlength{\tabcolsep}{4.0pt}
\begin{tabular}{@{}ll cc cc@{}}
\toprule
& & \multicolumn{2}{c}{\textbf{Qwen Judge}}
& \multicolumn{2}{c}{\textbf{Gemma Judge}} \\
\cmidrule(lr){3-4} \cmidrule(l){5-6}
\textbf{Dataset} & \textbf{Surface}
& \textbf{Sup CBU/cap}\,$\uparrow$ & \textbf{Risk}\,$\downarrow$
& \textbf{Sup CBU/cap}\,$\uparrow$ & \textbf{Risk}\,$\downarrow$ \\
\midrule
CC12M & Ours & $14.60 \pm 0.06$ & $0.030 \pm 0.001$ & $13.82 \pm 0.06$ & $0.066 \pm 0.001$ \\
 & CC12M-LLaVA-NeXT & $9.84 \pm 0.05$ & $0.069 \pm 0.002$ & $9.44 \pm 0.05$ & $0.097 \pm 0.002$ \\
 & PixelProse & $10.73 \pm 0.05$ & $0.129 \pm 0.002$ & $10.20 \pm 0.05$ & $0.161 \pm 0.002$ \\
 & CC12M-Qwen3-VL${}^{\dagger}$ & $6.31 \pm 0.03$ & $0.014 \pm 0.001$ & $6.19 \pm 0.03$ & $0.028 \pm 0.001$ \\
\addlinespace[0.2em]
DataComp & Reference & $8.49 \pm 0.05$ & $0.177 \pm 0.003$ & $8.00 \pm 0.05$ & $0.219 \pm 0.003$ \\
 & Ours & $13.73 \pm 0.05$ & $0.035 \pm 0.001$ & $12.94 \pm 0.05$ & $0.081 \pm 0.001$ \\
\addlinespace[0.2em]
LAION-pop & Reference & $10.80 \pm 0.04$ & $0.077 \pm 0.001$ & $10.22 \pm 0.04$ & $0.113 \pm 0.002$ \\
 & Ours & $14.22 \pm 0.05$ & $0.031 \pm 0.001$ & $13.61 \pm 0.05$ & $0.060 \pm 0.001$ \\
\addlinespace[0.2em]
PD12M & Reference & $8.61 \pm 0.05$ & $0.103 \pm 0.002$ & $8.23 \pm 0.05$ & $0.131 \pm 0.002$ \\
 & Ours & $14.29 \pm 0.05$ & $0.034 \pm 0.001$ & $13.53 \pm 0.05$ & $0.066 \pm 0.001$ \\
\addlinespace[0.2em]
Danbooru & Reference & $6.38 \pm 0.04$ & $0.217 \pm 0.003$ & $6.15 \pm 0.04$ & $0.235 \pm 0.003$ \\
 & Ours & $12.74 \pm 0.05$ & $0.058 \pm 0.001$ & $11.85 \pm 0.04$ & $0.094 \pm 0.001$ \\
\addlinespace[0.2em]
CC12M control & Naive & $11.03 \pm 0.05$ & $0.022 \pm 0.001$ & $10.61 \pm 0.05$ & $0.045 \pm 0.001$ \\
 & Naive (greedy) & $11.10 \pm 0.05$ & $0.022 \pm 0.001$ & $10.79 \pm 0.05$ & $0.043 \pm 0.001$ \\
\addlinespace[0.2em]
DataComp control & Ours & $13.90 \pm 0.06$ & $0.036 \pm 0.001$ & $13.07 \pm 0.05$ & $0.080 \pm 0.001$ \\
 & Naive & $10.56 \pm 0.04$ & $0.026 \pm 0.001$ & $10.07 \pm 0.04$ & $0.057 \pm 0.001$ \\
 & Naive (greedy) & $10.53 \pm 0.04$ & $0.021 \pm 0.001$ & $10.11 \pm 0.04$ & $0.049 \pm 0.001$ \\
\bottomrule
\end{tabular}
\end{table}

\clearpage
\subsection{Human Verification Detail}
\label{app:human_cbu_details}

\begin{figure}[t]
  \centering
  \begin{subfigure}[t]{0.485\linewidth}
    \centering
    \includegraphics[width=\linewidth]{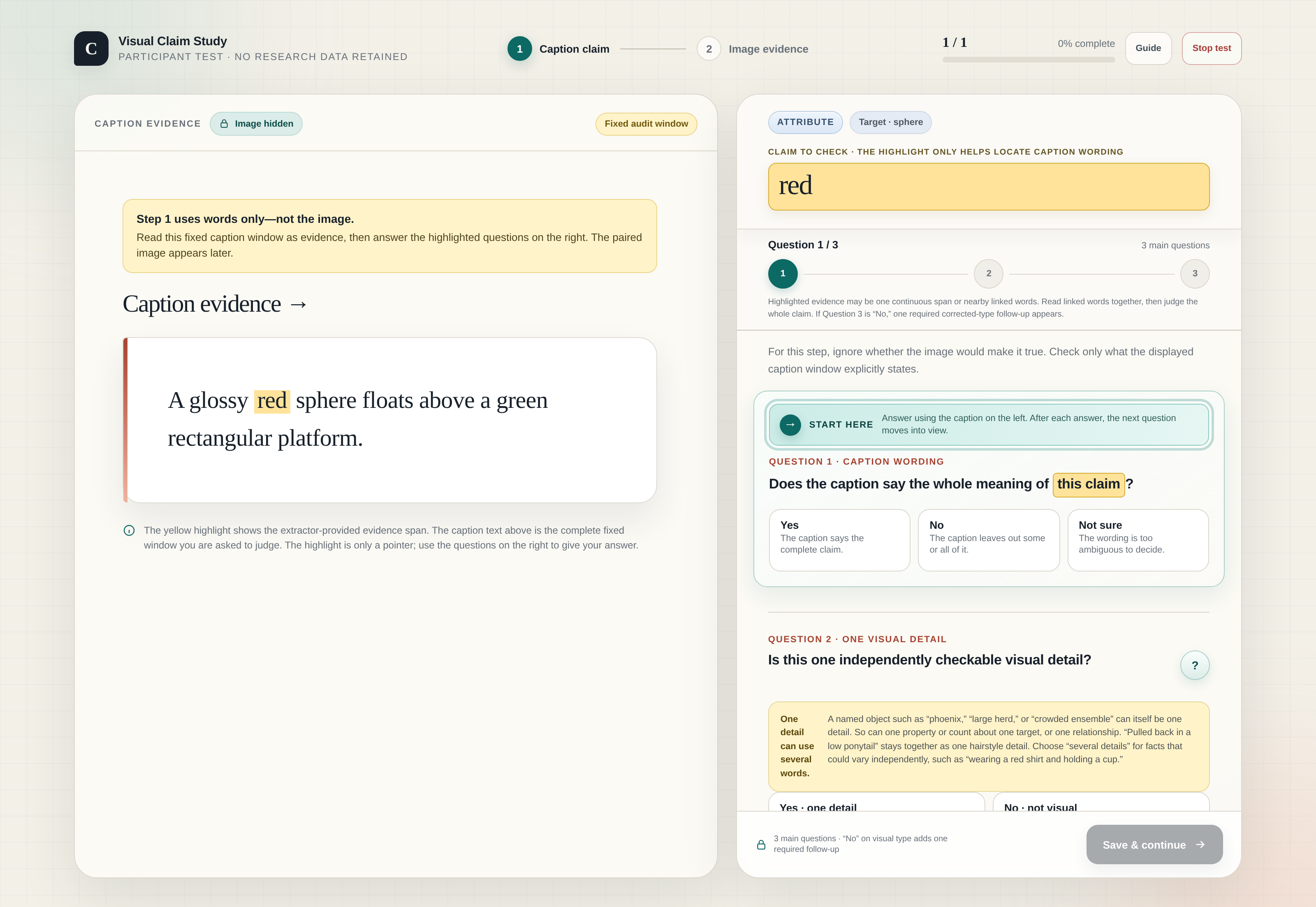}
    \caption{Caption judgment with the image hidden.}
  \end{subfigure}\hfill
  \begin{subfigure}[t]{0.485\linewidth}
    \centering
    \includegraphics[width=\linewidth]{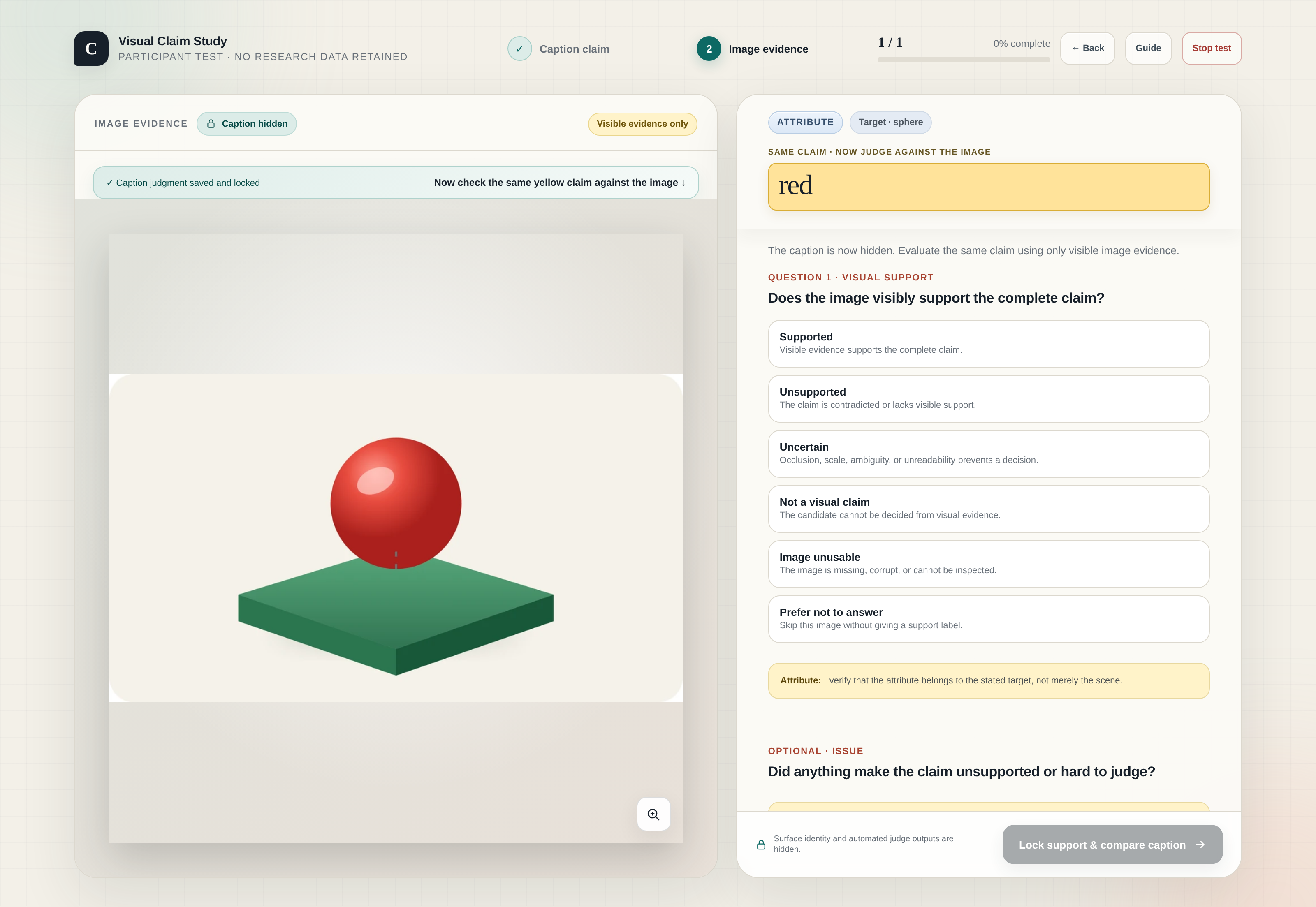}
    \caption{Image-support judgment with the caption hidden.}
  \end{subfigure}
  \caption{\textbf{Annotation interface.} Annotators first judge
  whether the fixed caption window licenses the atomic claim and its proposed
  semantic type, then judge the same claim against the image with the caption
  hidden. Surface identity, automated judgments, and other annotators'
  responses stay hidden throughout; the captures use a synthetic practice item.}
  \label{fig:human_cbu_protocol}
\end{figure}

Figure~\ref{fig:human_cbu_protocol} shows the annotation interface, and
Tables~\ref{tab:human_cbu_full_surface}--\ref{tab:human_cbu_by_type} carry the
complete human-verification grid behind Table~\ref{tab:human_cbu_descriptive}.
The surface-level census reports every caption-side screen and every image
response category, and the claim-type grid places the broadest Ours
margins in object and text-rendering claims.

\paragraph{Annotation setup.}
Seven volunteer annotators verified the sampled claims without compensation.
The annotation interface stores each judgment under an anonymous code, and the code repository carries the interface with its complete instruction sequence.

\begin{table}[h]
\centering
\caption{\textbf{Complete human-verification census by CC12M surface.}
$n$ counts primary non-repeat judgments and the remaining cells are unweighted
percentages; each surface has the same $n$ in the caption and image phases. ``Other'' combines
\emph{not visual} and \emph{prefer not to answer}. The four image columns
partition the phase denominator.}
\label{tab:human_cbu_full_surface}
\scriptsize
\renewcommand{\arraystretch}{1.08}
\setlength{\tabcolsep}{3.0pt}
\begin{tabular}{@{}lrrrrrrrr@{}}
\toprule
& & \multicolumn{3}{c}{\textbf{Caption screen}} &
\multicolumn{4}{c}{\textbf{Image support}} \\
\cmidrule(lr){3-5}\cmidrule(l){6-9}
\textbf{Surface} & \textbf{$n$} & \textbf{Licensed} & \textbf{Atomic} &
\textbf{Type} & \textbf{Yes} & \textbf{Uncertain} & \textbf{No} &
\textbf{Other} \\
\midrule
\textbf{Ours} & $111$ & $95.5$ & $86.5$ & $86.5$ &
$\mathbf{78.4}$ & $19.8$ & $\mathbf{0.0}$ & $1.8$ \\
LLaVA-NeXT & $47$ & $93.6$ & $91.5$ & $91.5$ &
$70.2$ & $21.3$ & $4.3$ & $4.3$ \\
Qwen3-VL-8B & $20$ & $95.0$ & $90.0$ & $90.0$ &
$70.0$ & $25.0$ & $5.0$ & $0.0$ \\
PixelProse & $39$ & $94.9$ & $84.6$ & $84.6$ &
$59.0$ & $23.1$ & $15.4$ & $2.6$ \\
\bottomrule
\end{tabular}
\end{table}

\begin{table}[h]
\centering
\caption{\textbf{Human image-support ``yes'' judgments by claim type.}
Cells are count / primary judgments (percentage) for the full eight-type grid.
Denominators follow the sampled claims, and a dash marks a surface--type
cell with no sampled claim.}
\label{tab:human_cbu_by_type}
\scriptsize
\renewcommand{\arraystretch}{1.08}
\setlength{\tabcolsep}{4.0pt}
\begin{tabular}{@{}lrrrr@{}}
\toprule
\textbf{Claim type} & \textbf{Ours} & \textbf{LLaVA-NeXT} &
\textbf{Qwen3-VL-8B} & \textbf{PixelProse} \\
\midrule
Attribute      & $8/12\;(66.7\%)$  & $4/8\;(50.0\%)$  & -- & $4/6\;(66.7\%)$ \\
Camera         & $16/16\;(100\%)$  & $11/14\;(78.6\%)$ & $1/1\;(100\%)$ & $2/2\;(100\%)$ \\
Count          & $10/12\;(83.3\%)$ & $1/1\;(100\%)$  & $2/2\;(100\%)$ & $5/10\;(50.0\%)$ \\
Lighting       & $6/13\;(46.2\%)$  & $4/6\;(66.7\%)$ & $5/6\;(83.3\%)$ & -- \\
Object         & $12/15\;(80.0\%)$ & $1/2\;(50.0\%)$ & $3/5\;(60.0\%)$ & $3/6\;(50.0\%)$ \\
Relation       & $11/14\;(78.6\%)$ & $2/2\;(100\%)$  & $1/3\;(33.3\%)$ & $5/7\;(71.4\%)$ \\
Style          & $14/16\;(87.5\%)$ & $6/6\;(100\%)$  & $2/2\;(100\%)$ & $3/3\;(100\%)$ \\
Text rendering & $10/13\;(76.9\%)$ & $4/8\;(50.0\%)$ & $0/1\;(0.0\%)$ & $1/5\;(20.0\%)$ \\
\bottomrule
\end{tabular}
\end{table}

\section{Extended Captioner-Control Results}
\label{app:naive_qwen35_baseline}

Phrase-presence rates and frequent raw prefixes describe the descriptive frames of the naive CC12M captions and complement Tables~\ref{tab:captioner_control_cbu}--\ref{tab:captioner_control_surface}.

\paragraph{Decoding and denominators.}
The naive surface decodes with the same release sampling defaults as Ours (Appendix~\ref{app:vlm_templates}).
Table~\ref{tab:policy_control_denominators} lists the audited captions and CBU questions behind Table~\ref{tab:captioner_control_cbu}.
The DataComp control runs on $4{,}775$ images and the cross-corpus DataComp pair on $4{,}922$, so the two Ours rows carry their own denominators.
Greedy decoding of the naive prompt yields $11.43$ and $10.84$ claimed CBU per caption on CC12M and DataComp.

\begin{table}[h]
\centering
\caption{\textbf{Captioner-control denominators at $B = 64$.} VQA: audited captions; Q: CBU questions, identical for both Judges. Naive (greedy) decodes the naive prompt at temperature $0$.}
\label{tab:policy_control_denominators}
\footnotesize
\renewcommand{\arraystretch}{1.08}
\setlength{\tabcolsep}{4.0pt}
\begin{tabular}{@{}ll rr cc cc@{}}
\toprule
& & & & \multicolumn{2}{c}{\textbf{Qwen Judge}} & \multicolumn{2}{c}{\textbf{Gemma Judge}} \\
\cmidrule(lr){5-6} \cmidrule(l){7-8}
\textbf{Dataset} & \textbf{Surface} & \textbf{VQA} & \textbf{Q} & \textbf{Sup CBU/cap}\,$\uparrow$ & \textbf{Risk}\,$\downarrow$ & \textbf{Sup CBU/cap}\,$\uparrow$ & \textbf{Risk}\,$\downarrow$ \\
\midrule
CC12M & Ours & $4{,}467$ & $68{,}019$ & $14.60$ & $0.030$ & $13.82$ & $0.066$ \\
 & Naive & $4{,}397$ & $50{,}450$ & $11.03$ & $0.022$ & $10.61$ & $0.045$ \\
 & Naive (greedy) & $4{,}423$ & $51{,}365$ & $11.10$ & $0.022$ & $10.79$ & $0.043$ \\
\addlinespace[0.2em]
DataComp & Ours & $4{,}688$ & $68{,}473$ & $13.90$ & $0.036$ & $13.07$ & $0.080$ \\
 & Naive & $4{,}618$ & $50{,}853$ & $10.56$ & $0.026$ & $10.07$ & $0.057$ \\
 & Naive (greedy) & $4{,}607$ & $50{,}353$ & $10.53$ & $0.021$ & $10.11$ & $0.049$ \\
\bottomrule
\end{tabular}
\end{table}

\paragraph{Phrase presence and raw prefixes.}
The naive surface opens predominantly with \texttt{This is a [adjective]\dots}, a frame outside the opener-rate regex of Section~\ref{sec:prelim:gap}, and two complementary signals register it.
Caption-level phrase-presence rates: \texttt{this is} $92.6\%$, \texttt{overall} $85.5\%$, \texttt{the image} $57.1\%$, \texttt{likely} $55.8\%$, \texttt{appears to} $39.8\%$, \texttt{the scene} $40.3\%$, \texttt{vibrant} $38.8\%$.
Five most frequent 5-token raw prefixes: \texttt{``this is a close up''} ($238$$/4{,}494$ captions), \texttt{``this is a richly detailed''} ($157$), \texttt{``this is a full length''} ($152$), \texttt{``this is a beautifully composed''} ($121$), and \texttt{``this is a wide angle''} ($92$). Top-100 raw prefix mass is $0.60$ on the naive surface, more than $4{\times}$ Ours ($0.14$) and close to PixelProse ($0.71$), whose instruction hard-codes its opener.

\section{Sensitivity Diagnostics}
\label{app:sensitivity}

\begin{table}[h]
\centering
\caption{\textbf{DataComp text-space rankings vary by encoder and metric.}
50k paired rows. Vendi~\citep{friedman2023vendi} = exponential of the eigenvalue entropy of the caption-set kernel; eRank~\citep{roy2007erank} = significant directions in the centred embedding covariance. Prompt-to-caption PRDC~\citep{naeem2020prdc} uses DiffusionDB~\citep{wang2023diffusiondb} prompts at $k = 10$. kNN cos $=$ caption-to-caption directional cosine at $k = 10$, separately for Ours-as-query ($\rightarrow$) and Ref-as-query ($\leftarrow$).}
\label{tab:datacomp_text_audits}
\footnotesize
\renewcommand{\arraystretch}{1.08}
\setlength{\tabcolsep}{3.0pt}
\begin{tabular}{@{}l rr rr rr rr rr@{}}
\toprule
& \multicolumn{2}{c}{\textbf{Vendi}$\,\uparrow$}
& \multicolumn{2}{c}{\textbf{eRank}$\,\uparrow$}
& \multicolumn{2}{c}{\textbf{Coverage@10}$\,\uparrow$}
& \multicolumn{2}{c}{\textbf{Density@10}$\,\uparrow$}
& \multicolumn{2}{c}{\textbf{kNN cos}$\,\sim$} \\
\cmidrule(lr){2-3}
\cmidrule(lr){4-5}
\cmidrule(lr){6-7}
\cmidrule(lr){8-9}
\cmidrule(l){10-11}
\textbf{Encoder}
& \textbf{Ours} & \textbf{Ref}
& \textbf{Ours} & \textbf{Ref}
& \textbf{Ours} & \textbf{Ref}
& \textbf{Ours} & \textbf{Ref}
& $\boldsymbol{\rightarrow}$ & $\boldsymbol{\leftarrow}$ \\
\midrule
Qwen3-Emb-4B
& $61.40$ & $\mathbf{62.57}$
& $\mathbf{337.2}$ & $319.5$
& $\mathbf{0.743}$ & $0.357$
& $\mathbf{0.218}$ & $0.066$
& $0.758$ & $0.760$ \\

Qwen3-Emb-8B
& $76.33$ & $\mathbf{89.25}$
& $\mathbf{424.0}$ & $378.5$
& -- & --
& -- & --
& $0.721$ & $0.721$ \\

BGE-M3
& $25.39$ & $\mathbf{34.21}$
& $\mathbf{255.1}$ & $235.5$
& $0.017$ & $\mathbf{0.050}$
& $0.002$ & $\mathbf{0.007}$
& $0.776$ & $0.776$ \\
\bottomrule
\end{tabular}
\end{table}

\paragraph{LongCLIP retrieval, full-caption mode.}
Table~\ref{tab:cc12m_longclip} reports full-caption LongCLIP retrieval on the same $4{,}494$-image CC12M subset. Ours leads R@1 by a small margin in both directions, with $32.4\%$ of its captions over the $248$-token cap; the naive surface of Section~\ref{sec:captioner_control}, with $99.2\%$ of captions over the cap, scores $0.906$ I2T R@1 in full mode and $0.812$ in input64 mode, where it is the lowest of the five surfaces (Table~\ref{tab:cc12m_longclip_input64}).

\begin{table}[h]
\centering
\caption{LongCLIP retrieval on CC12M (full-caption mode). \texttt{LongCLIP-GmP-ViT-L-14}~\citep{longclip2024}, $4{,}494$-image subset, retrieval pool $4{,}494$; \emph{trunc.} is the caption rate exceeding the $248$-token cap. Pre-truncated input64 mode in Table~\ref{tab:cc12m_longclip_input64}; dagger ${}^{\dagger}$ marks short tag-style surfaces (Section~\ref{sec:prelim:targets}).}
\label{tab:cc12m_longclip}
\small
\renewcommand{\arraystretch}{1.10}
\setlength{\tabcolsep}{4pt}
\begin{tabular}{@{}l rrr rr rr@{}}
\toprule
& \multicolumn{3}{c}{LongCLIP tokens / cap}
& \multicolumn{2}{c}{I2T$\,\uparrow$}
& \multicolumn{2}{c}{T2I$\,\uparrow$} \\
\cmidrule(lr){2-4} \cmidrule(lr){5-6} \cmidrule(l){7-8}
\textbf{Surface} & mean & p95 & trunc. & R@1 & R@5 & R@1 & R@5 \\
\midrule
Ours
  & $231.7$ & $320$ & $32.4\%$ & $\mathbf{0.908}$ & $0.988$ & $\mathbf{0.902}$ & $\mathbf{0.986}$ \\

CC12M-LLaVA-NeXT~\citep{conceptualcaptionscc12mllavanext}
  & $81.9$  & $153$ & $0.4\%$  & $0.907$ & $\mathbf{0.989}$ & $0.899$ & $\mathbf{0.986}$ \\

PixelProse~\citep{singla2024pixelprose}
  & $108.4$ & $193$ & $1.7\%$  & $0.890$ & $0.977$ & $0.874$ & $0.976$ \\

CC12M-Qwen3-VL${}^{\dagger}$~\citep{cc12mQwen3VL8BHF}
  & $15.4$  & $20$  & $0\%$    & $0.861$ & $0.979$ & $0.842$ & $0.971$ \\

Naive Qwen3.5-35B-A3B
  & $424.5$ & $522$ & $99.2\%$ & $0.906$ & $0.984$ & $0.882$ & $0.977$ \\
\bottomrule
\end{tabular}
\end{table}

\paragraph{LongCLIP retrieval, input64 mode.}
Table~\ref{tab:cc12m_longclip_input64} mirrors Table~\ref{tab:cc12m_longclip} after pre-truncating each caption to $B = 64$ lexical units before LongCLIP encoding. CC12M-LLaVA-NeXT moves to the highest R@1 in both retrieval directions, PixelProse follows, and Ours places third on T2I ($0.850$) and within $0.002$ of CC12M-Qwen3-VL${}^{\dagger}$ on I2T ($0.859$ against $0.861$), the reverse of its full-mode lead.

\begin{table}[h]
\centering
\caption{LongCLIP retrieval on CC12M (input64 mode): captions pre-truncated to $B = 64$ lexical units before LongCLIP encoding. Same $4{,}494$-image subset and column structure as Table~\ref{tab:cc12m_longclip}.}
\label{tab:cc12m_longclip_input64}
\small
\renewcommand{\arraystretch}{1.10}
\setlength{\tabcolsep}{4pt}
\begin{tabular}{@{}l rrr rr rr@{}}
\toprule
& \multicolumn{3}{c}{LongCLIP tokens / cap}
& \multicolumn{2}{c}{I2T$\,\uparrow$}
& \multicolumn{2}{c}{T2I$\,\uparrow$} \\
\cmidrule(lr){2-4} \cmidrule(lr){5-6} \cmidrule(l){7-8}
\textbf{Surface} & mean & p95 & trunc. & R@1 & R@5 & R@1 & R@5 \\
\midrule
Ours                          & $81.8$ & $90$ & $0\%$    & $0.859$         & $0.971$         & $0.850$         & $0.965$ \\
CC12M-LLaVA-NeXT              & $60.6$ & $83$ & $0\%$    & $\mathbf{0.897}$ & $\mathbf{0.985}$ & $\mathbf{0.888}$ & $\mathbf{0.983}$ \\
PixelProse                    & $73.4$ & $82$ & $0.07\%$ & $0.869$         & $0.968$         & $0.856$         & $0.963$ \\
CC12M-Qwen3-VL${}^{\dagger}$  & $15.4$ & $20$ & $0\%$    & $0.861$         & $0.979$         & $0.842$         & $0.971$ \\
Naive Qwen3.5-35B-A3B       & $82.6$ & $91$ & $0\%$ & $0.812$ & $0.948$ & $0.789$ & $0.934$ \\
\bottomrule
\end{tabular}
\end{table}

\paragraph{Encoder-token truncation, full per-surface rates.}
Table~\ref{tab:encoder_truncation} reports per-encoder truncation rate (the share of captions exceeding the encoder's token cap) on a 100k-row sample per surface ($4{,}494$ captions for the naive control). Long Ours captions exceed CLIP-77 and SigLIP2-64 on $\sim 100\%$ of captions and LongCLIP-248 on $22$--$40\%$; short comparison surfaces (CC12M-Qwen3-VL${}^{\dagger}$, PD12M released) fit most encoders.

\begin{table}[h]
\centering
\caption{Encoder-token truncation rate on three image-text contrastive alignment encoders --- CLIP-77~\citep{radford2021clip}, LongCLIP-248~\citep{longclip2024}, SigLIP2-64~\citep{tschannen2025siglip2} --- measured on 100k-row samples per surface ($4{,}494$ captions for the naive control). Numbers are the fraction of captions exceeding the encoder's token cap; mean tokens count CLIP tokens.}
\label{tab:encoder_truncation}
\footnotesize
\renewcommand{\arraystretch}{1.10}
\setlength{\tabcolsep}{4pt}
\begin{tabular}{l l r r r r}
\toprule
\textbf{Source} & \textbf{Surface} & \textbf{Mean tokens} & \textbf{CLIP-77 trunc} & \textbf{LongCLIP-248 trunc} & \textbf{SigLIP2-64 trunc} \\
\midrule
DataComp   & Ours                       & $223$ & $99.98\%$  & $25.41\%$ & $99.99\%$  \\
           & Recap-DataComp            & $62$  & $25.21\%$  & $0.01\%$  & $41.69\%$  \\
\addlinespace
CC12M      & Ours                       & $231$ & $99.97\%$  & $31.24\%$ & $99.98\%$  \\
           & Naive Qwen3.5-35B-A3B         & $425$ & $100.00\%$ & $99.24\%$ & $100.00\%$ \\
           & CC12M-LLaVA-NeXT           & $88$  & $57.61\%$  & $0.49\%$  & $64.09\%$  \\
           & PixelProse                 & $108$ & ---       & $2.23\%$  & ---       \\
           & CC12M-Qwen3-VL${}^{\dagger}$ & $15$  & $0.00\%$   & $0.00\%$  & $0.00\%$   \\
\addlinespace
LAION-pop  & Ours                       & $228$ & $99.99\%$  & $29.12\%$ & $99.99\%$  \\
           & LAION-pop-Llama            & $218$ & $99.98\%$  & $25.32\%$ & $100.00\%$ \\
\addlinespace
PD12M      & Ours                       & $247$ & $99.99\%$  & $40.56\%$ & $99.99\%$  \\
           & PD12M released             & $47$  & $5.43\%$   & $0.00\%$  & $12.79\%$  \\
\addlinespace
Danbooru   & Ours                       & $219$ & $99.89\%$  & $22.27\%$ & $99.95\%$  \\
           & Danbooru-Florence          & $51$  & $12.88\%$  & $0.00\%$  & $25.59\%$  \\
\bottomrule
\end{tabular}
\end{table}

\paragraph{Embedding-protocol details.}
Caption embeddings under \texttt{Qwen3-Embedding-4B} and \texttt{Qwen3-Embedding-8B} use last-token pooling with left padding, $\mathrm{max\_length}=512$, and L2 normalization, following the official Qwen3-Embedding usage card; \texttt{BGE-M3} uses \texttt{FlagEmbedding.BGEM3FlagModel} dense vectors; \texttt{BGE-large-en-v1.5} uses CLS pooling; \texttt{E5-Mistral-7B-Instruct} uses last-token pooling with $\mathrm{max\_length}=4096$.
The PRDC columns of Table~\ref{tab:datacomp_text_audits} report the raw-text protocol for each encoder (no query-side instruction prefix); the result summaries also hold a query/document instruction protocol for E5-Mistral and Qwen3-Embedding-4B (Appendix~\ref{app:release_manifest}).

\section{Release Artifacts, Hosting, and Croissant/RAI Coverage}
\label{app:release_manifest}

\paragraph{Caption record schema.}
The release is one caption-only dataset per source family, and Table~\ref{tab:release_schema} groups the fields of a caption row.
Source keys locate the image in its source release, the image locator binds the row to the image package of that family, the caption fields hold the released text with its hash and record identifier, and the generation fields record the captioner artifact and token counts.
DataComp rows carry a stable image identifier with the normalized source URL, and Megalith-CC0 rows carry the source repository revision and native identifier.

\begin{table}[h]
\centering
\caption{\textbf{Caption record fields of the public release.}}
\label{tab:release_schema}
\footnotesize
\renewcommand{\arraystretch}{1.10}
\setlength{\tabcolsep}{4.0pt}
\begin{tabularx}{\linewidth}{@{}l >{\raggedright\arraybackslash}X l@{}}
\toprule
\textbf{Group} & \textbf{Fields} & \textbf{Role} \\
\midrule
Source keys & \texttt{source\_dataset}, \texttt{source\_url}, \texttt{source\_url\_norm}, \texttt{source\_url\_sha1}, \texttt{source\_sha256} & source join \\
Image locator & \texttt{asset\_instance\_id}, \texttt{image\_shard}, \texttt{image\_member} & image binding \\
Caption & \texttt{caption\_text}, \texttt{caption\_sha256}, \texttt{caption\_record\_id} & released text \\
Generation & \texttt{caption\_model\_family}, \texttt{caption\_model\_artifact}, \texttt{prompt\_tokens}, \texttt{completion\_tokens} & provenance \\
\bottomrule
\end{tabularx}
\end{table}

\paragraph{Reproduction workflow.}
The public collection carries our generated captions for all nine source families; the code repository carries the audit pipeline, the CC12M and DataComp naive-control surfaces, and the result summaries behind every table.
Reference recaptions and source images are reached through their public releases and joined by URL, UUID, SHA-256, or dataset-native keys; Danbooru images sit in a separate gated repository.
Text statistics and claim extraction run on the released captions alone, and image-conditioned verification adds paired image access and the Judge schema of Appendix~\ref{app:vlm_templates}.

\paragraph{Personal information and training use.}
Captions are released as generated: the captioner transcribes legible text, so a caption can carry a name, a contact detail, or an identifier that the source image shows.
The text stays unmodified so that each user applies the filter that fits their use; every dataset card documents this and routes removal requests through the repository's discussion page with the row keys.
The naive-control captions in the code repository mask such strings with placeholders, and every family is research data for which training pipelines apply their own safety and policy filters.

\paragraph{Result provenance.}
Table~\ref{tab:artifact_provenance} lists the released artifact behind each main measurement.

\begin{table}[h]
\centering
\small
\caption{\textbf{Result provenance.} Claim and VQA summaries use $B = 64$, and both Judges supply the VQA answers.}
\label{tab:artifact_provenance}
\renewcommand{\arraystretch}{1.10}
\begin{tabularx}{\linewidth}{@{}l l X@{}}
\toprule
\textbf{Result} & \textbf{Scope} & \textbf{Artifact} \\
\midrule
Claim yield (Table~\ref{tab:cross_full}) & cross-corpus pairs & \path{all_cbu_b64_summary.csv} \\
Support and risk (Tables~\ref{tab:cross_full}, \ref{tab:cc12m_frontier}, Fig.~\ref{fig:cc12m_frontiers}) & both Judges & \path{cbu_vqa_by_category_b64.json} \\
Length and prefix statistics (Tables~\ref{tab:phenomenon_meta_opener}, \ref{tab:cross_full}) & caption text & \path{cpu_text_metrics/} \\
Pool comparisons (Figs.~\ref{fig:teaser}, \ref{fig:per_pool_heatmap}) & seven prompt pools & \path{prompt_support_bootstrap_b64_n2_250k_2026-04-24.tsv} \\
Claim yield and budget sweep (Table~\ref{tab:cc12m_frontier}, Fig.~\ref{fig:cc12m_frontiers}) & CC12M & \path{cc12m_budget_frontier_plot.csv} \\
Policy control (Tables~\ref{tab:captioner_control_cbu}, \ref{tab:captioner_control_surface}) & CC12M, DataComp & \path{naive_qwen35_*/} \\
Release metadata (Section~\ref{sec:release}) & Croissant core + RAI & \path{croissant.json} per dataset \\
\bottomrule
\end{tabularx}
\end{table}

\paragraph{Hosting and Croissant coverage.}
Each dataset of the collection carries a Croissant~1.1 record~\citep{akhtar2024croissant,jain2024croissantrai}, \texttt{croissant.json} at the repository root, with the core fields (\texttt{name}, \texttt{description}, \texttt{license}, \texttt{citeAs}, \texttt{distribution}, \texttt{recordSet}) and the Responsible-AI fields (\texttt{rai:dataLimitations}, \texttt{rai:dataBiases}, \texttt{rai:personalSensitiveInformation}, \texttt{rai:dataUseCases}, \texttt{rai:dataSocialImpact}, \texttt{rai:hasSyntheticData}, \texttt{prov:wasDerivedFrom}, \texttt{prov:wasGeneratedBy}).
Every record passes the \texttt{mlcroissant} validator, and the Hub serves a core record generated from the data files at each dataset's Croissant endpoint; code and result summaries are at \url{https://github.com/BootsofLagrangian/audit-recap-t2i}.

\section{VLM Request Templates}
\label{app:vlm_templates}

\paragraph{Checkpoints.}
$V_c$: \texttt{Qwen3.5-35B-A3B-FP8}~\footnote{\url{https://huggingface.co/Qwen/Qwen3.5-35B-A3B-FP8}}.
Claim extractor $\phi$ and Qwen Judge $V_J$: \texttt{Qwen3.5-397B-A17B-FP8}~\footnote{\url{https://huggingface.co/Qwen/Qwen3.5-397B-A17B-FP8}}.
Gemma Judge: \texttt{Gemma-4-31B-IT}~\footnote{\url{https://huggingface.co/google/gemma-4-31B-it}}.

All VLM stages use OpenAI-compatible chat-completion endpoints served by vLLM~\citep{kwon2023vllm}. Caption generation (Ours and the Naive ablation of Appendix~\ref{app:naive_qwen35_baseline}) uses the Qwen3.5 release sampling defaults~\citep{qwen35} with free-form text output. The Qwen claim extractor $\phi$ and the Qwen and Gemma Judges use deterministic decoding ($\mathrm{temperature}=0$) and schema-constrained JSON via vLLM guided decoding.
The request templates below are the verbatim VLM system and user prompts used in the evaluation: photorealistic and anime caption generation, the captioner-control naive prompt, the claim extractor $\phi$, and the image-conditioned Judge. The same templates ship with the released audit-artifact bundle.

\paragraph{Caption generation, photorealistic domain.}
\noindent\emph{System prompt.}
\begin{lstlisting}
You write photorealistic training captions for image generation models.
Start with the dominant subject or action, then expand through clothing
or materials, props, environment, lighting, atmosphere, and camera
framing. Describe the scene directly in present-tense prose as if
narrating what is there rather than commenting on a photograph.
Describe people by visible attributes only.
If the image contains large, clearly legible text, append a TEXT: block
with the unique lines only. Do not repeat the same text more than once
even if it appears multiple times in the image. Omit TEXT: for partial,
small, or ambiguous text. Write factual, visually specific prose with
no headers, bullets, lists, markdown, or pipeline references.
\end{lstlisting}
\noindent\emph{User prompt} (single long-caption stage, $\mathrm{max\_tokens}=512$).
\begin{lstlisting}
Write one detailed photorealistic caption.
Start with the dominant subject or action, then expand through scene
layout, surfaces and materials, lighting, atmosphere, and camera
composition. Keep the prose descriptive, grounded, and specific.
\end{lstlisting}

\paragraph{Caption generation, anime / booru domain.}
\noindent\emph{System prompt} (booru-tag-grounded).
\begin{lstlisting}
You write anime training captions grounded by booru metadata.
Start with the dominant subject or action, then expand outward through
anatomy, clothing, props, other figures, environment, lighting, and
framing. Treat provided booru tags as grounding anchors and weave them
into natural prose. When confirmed_characters are provided, always open
the caption using a name from the list. The list is ordered from the
broadest base identity to the most specific costume or variant. The
last entry is typically the most precise match for what is depicted.
All entries for a solo image refer to the same individual at different
levels of specificity. You must use at least one name; prefer the most
specific entry that matches the visible appearance. If no
confirmed_characters are provided, describe the figure by visible
attributes only. If the image contains large, clearly legible text,
append a TEXT: block with the unique lines only. Do not repeat the
same text more than once even if it appears multiple times in the
image. Omit TEXT: for partial, small, or ambiguous text. Write
factual, present-tense, visually specific prose with no headers,
bullets, lists, markdown, or pipeline references.
\end{lstlisting}
\noindent\emph{User prompt} (single long-caption stage, $\mathrm{max\_tokens}=512$, with the ordered booru context fields \texttt{general\_tags}, \texttt{character\_tags}, \texttt{confirmed\_characters}, \texttt{copyright\_tags}, \texttt{rating} provided alongside the image).
\begin{lstlisting}
Write one detailed anime training caption grounded by the image and
booru context. Start with the dominant subject or action, then expand
outward through anatomy, attire, props, setting, mood, style, lighting,
and composition. When confirmed character names are provided, open with
a name from the list. Names sharing a common root are the same person
at increasing levels of costume or form detail --- always pick the most
detailed variant whose outfit or form matches the image. Names with
entirely different roots in a solo image are contextual references to
a partially visible or implied second character; only name the visually
dominant one. If no confirmed names are given, describe the figure
visually. Deliver continuous prose, and append a TEXT: block with
unique lines only when text is clearly readable. Do not repeat the
same text more than once.
\end{lstlisting}

\paragraph{Captioner-control ablation (naive prompt).}
The captioner-control ablation in Section~\ref{sec:captioner_control} reuses the same captioner $V_c = \mathtt{Qwen3.5\text{-}35B\text{-}A3B\text{-}FP8}$ under a single user message and no system prompt; the prompt body inherits from the Recap-DataComp instruction.
\noindent\emph{User prompt} (no system prompt, $\mathrm{max\_tokens}=512$).
\begin{lstlisting}
Please generate a detailed caption of this image. Please be as
descriptive as possible.
\end{lstlisting}

\paragraph{Claim extractor $\phi$ (text-only, Qwen3.5-397B-A17B-FP8).}
\noindent\emph{System prompt.}
\begin{lstlisting}
You extract atomic controllable visual content units from captions for
text-to-image training-data evaluation. Return only valid compact JSON.
Extract only facts explicitly claimed by the caption. Do not infer
image content beyond the caption.
\end{lstlisting}
\noindent\emph{User prompt} (\texttt{<CAPTION\_ID>}, \texttt{<CAPTION\_TEXT\_B>} are the caption identifier and the first $B$ whitespace-delimited words of the caption; the JSON schema constrains the output to the eight CBU categories listed in Section~\ref{sec:protocol:axes}).
\begin{lstlisting}
Extract caption-claimed controllable visual units as atomic records.
Unit categories: object, attribute, relation, style, camera, lighting,
count, text_rendering.
Rules:
- Each record must contain exactly one visual control fact.
- Use each semantic fact once; choose the single best category.
- unit is a short canonical phrase, not a full clause.
- span is the shortest caption span supporting the unit.
- target is the object or scene element modified by the unit;
  use "scene" when global.
- relation units must include both the relation and participating
  objects; do not output lone verbs or prepositions.
- count units must attach a number to a target object; never output
  articles such as a, an, or the.
- text_rendering units are only visible rendered text explicitly
  claimed by the caption; absent text claims are not units.
- Do not output negative or absent facts, metadata, captioner phrases,
  or duplicate paraphrases.
- Keep text_rendering units short; do not copy long copyright, table,
  or legal text blocks.
- Use [] when the caption contains no controllable visual units.
Return only JSON matching this schema:
<CLAIMED_CBU_JSON_SCHEMA>

caption_id=<CAPTION_ID>
caption=<CAPTION_TEXT_B>
\end{lstlisting}
\noindent\emph{Guided JSON schema} (the placeholder \texttt{<CLAIMED\_CBU\_JSON\_SCHEMA>} above is filled in via vLLM guided decoding with the schema below; the eight \texttt{category} values match the CBU categories of Section~\ref{sec:protocol:axes}).
\begin{lstlisting}
{
  "type": "object",
  "properties": {
    "caption_id": {"type": "string"},
    "claimed_units": {
      "type": "array",
      "items": {
        "type": "object",
        "properties": {
          "category": {
            "type": "string",
            "enum": ["object", "attribute", "relation", "style",
                     "camera", "lighting", "count", "text_rendering"]
          },
          "unit":   {"type": "string", "maxLength": 80},
          "span":   {"type": "string", "maxLength": 120},
          "target": {"type": "string", "maxLength": 80}
        },
        "required": ["category", "unit", "span", "target"],
        "additionalProperties": false
      }
    }
  },
  "required": ["caption_id", "claimed_units"],
  "additionalProperties": false
}
\end{lstlisting}

\paragraph{Image-conditioned Judge $V_J$ (Qwen3.5-397B-A17B-FP8 and Gemma-4-31B-IT, cross-corpus and CC12M).}
\noindent\emph{Question construction.} Each extracted CBU becomes one of three deterministic question forms:
\begin{lstlisting}
Is the visual claim '<TARGET>: <UNIT>' supported by the image?
Is the visual claim '<UNIT>' supported by the image?
Is the rendered text claim '<UNIT>' visibly supported by the image?
\end{lstlisting}
\noindent\emph{System prompt.}
\begin{lstlisting}
You are a strict visual question answering judge.
Return only valid compact JSON. Answer each question using only
visible image evidence.
\end{lstlisting}
\noindent\emph{User prompt} (compact answer-only mode).
\begin{lstlisting}
Answer each visual question using only the image.
Rules:
- Do not use any caption text or outside knowledge.
- Use yes when the image visibly supports the question.
- Use no when the image contradicts the question or lacks visible
  support.
- Use uncertain when the question is too fine-grained, occluded,
  unreadable, or visually ambiguous.
- Return only question_id, answer, and confidence for each question;
  do not include evidence text.
- Return exactly one answer for each input question_id.

questions=<QUESTIONS_JSON>
\end{lstlisting}
The audit aggregates this stage as $\mathrm{support} = \mathrm{yes}/\mathrm{questions}$, $\mathrm{uncertainty} = \mathrm{uncertain}/\mathrm{questions}$, and $\mathrm{risk} = \mathrm{no}/\mathrm{questions}$; deterministic question construction fixes the question denominator.

\section{DataComp Crawl and Survival Snapshot}
\label{app:crawl_survival}

The full-corpus release (Section~\ref{sec:release}) draws on a multi-source crawl pipeline using \texttt{img2dataset} with a $\mathrm{min\_short\_edge}=256$ pre-decode constraint; this section reports DataComp, the largest single source family.
Two survival quantities are tracked per source family: \emph{download-stage survival} (the fraction of attempted URLs whose payload was successfully fetched) and \emph{end-to-end usable survival} (the fraction also passing decode and quality gates).

Table~\ref{tab:datacomp_survival} reports the DataComp forward-partition crawl snapshot; the deployed corpus combines forward and backward partitions to $325.6$M caption rows over $325.5$M unique normalized URLs (Section~\ref{sec:release}).

\begin{table}[h]
\centering
\small
\caption{DataComp forward-partition crawl-and-survival snapshot at \texttt{2026-04-03T15:03:36Z} cutoff. Crawling continued past the cutoff, and the released totals exceed this snapshot; the processed share is taken over target URLs and every other share over processed URLs. The dominant post-download loss is \texttt{image\_too\_small}, driven by the $\mathrm{min\_short\_edge}=256$ gate.}
\label{tab:datacomp_survival}
\begin{tabularx}{0.85\linewidth}{X r r}
\toprule
Stage & Count & Share \\
\midrule
Target URLs            & $940{,}890{,}257$ & ---     \\
Processed              & $371{,}813{,}492$ & $39.5\%$ \\
\addlinespace
Download-stage survival & $282{,}860{,}383$ & $76.1\%$ \\
Usable images & $147{,}035{,}107$ & $39.6\%$ \\
\addlinespace
\multicolumn{3}{l}{\emph{Top failure families:}} \\
\texttt{image\_too\_small}      & $129{,}765{,}982$ & $34.9\%$ \\
\texttt{http\_403}              & $24{,}009{,}304$  & $6.5\%$  \\
\texttt{http\_404}              & $23{,}598{,}377$  & $6.4\%$  \\
\texttt{dns\_name\_not\_known}  & $10{,}921{,}933$  & $2.9\%$  \\
\bottomrule
\end{tabularx}
\end{table}

\clearpage
\section*{NeurIPS Paper Checklist}

\begin{enumerate}

\item {\bf Claims}
    \item[] Question: Do the main claims made in the abstract and introduction accurately reflect the paper's contributions and scope?
    \item[] Answer: \answerYes{} % Replace by \answerYes{}, \answerNo{}, or \answerNA{}.
    \item[] Justification: The abstract and Section~\ref{sec:intro} state the audit framework, the paired comparisons, and the release, and Section~\ref{sec:limitations} states their scope.

\item {\bf Limitations}
    \item[] Question: Does the paper discuss the limitations of the work performed by the authors?
    \item[] Answer: \answerYes{} % Replace by \answerYes{}, \answerNo{}, or \answerNA{}.
    \item[] Justification: Section~\ref{sec:limitations} covers the measurement, Judge, and generalization scope of the audit.

\item {\bf Theory assumptions and proofs}
    \item[] Question: For each theoretical result, does the paper provide the full set of assumptions and a complete (and correct) proof?
    \item[] Answer: \answerNA{} % Replace by \answerYes{}, \answerNo{}, or \answerNA{}.
    \item[] Justification: We do not have any theoretical results.

    \item {\bf Experimental result reproducibility}
    \item[] Question: Does the paper fully disclose all the information needed to reproduce the main experimental results of the paper to the extent that it affects the main claims and/or conclusions of the paper (regardless of whether the code and data are provided or not)?
    \item[] Answer: \answerYes{} % Replace by \answerYes{}, \answerNo{}, or \answerNA{}.
    \item[] Justification: Appendices~\ref{app:protocol} and~\ref{app:vlm_templates} give the measurement procedure, checkpoints, request templates, JSON schemas, and decoding settings.

\item {\bf Open access to data and code}
    \item[] Question: Does the paper provide open access to the data and code, with sufficient instructions to faithfully reproduce the main experimental results, as described in supplemental material?
    \item[] Answer: \answerYes{} % Replace by \answerYes{}, \answerNo{}, or \answerNA{}.
    \item[] Justification: The audited corpus is a public Hugging Face collection with Croissant 1.1 and Responsible-AI metadata, and the code repository carries the audit pipeline and result summaries; Appendix~\ref{app:release_manifest} documents hosting, the record schema, and the reproduction workflow.

\item {\bf Experimental setting/details}
    \item[] Question: Does the paper specify all the training and test details (e.g., data splits, hyperparameters, how they were chosen, type of optimizer) necessary to understand the results?
    \item[] Answer: \answerYes{} % Replace by \answerYes{}, \answerNo{}, or \answerNA{}.
    \item[] Justification: Section~\ref{sec:eval} and Appendix~\ref{app:protocol} give the evaluation samples, the text budget, the prompt pools, and the decoding settings.

\item {\bf Experiment statistical significance}
    \item[] Question: Does the paper report error bars suitably and correctly defined or other appropriate information about the statistical significance of the experiments?
    \item[] Answer: \answerYes{} % Replace by \answerYes{}, \answerNo{}, or \answerNA{}.
    \item[] Justification: Appendix Table~\ref{tab:bootstrap_ci} reports supported yield and risk of Sections~\ref{sec:cross}--\ref{sec:cc12m} as mean $\pm$ caption-level bootstrap standard deviation, and Table~\ref{tab:human_cbu_descriptive} reports Judge--human agreement with image-cluster bootstrap standard deviations.

\item {\bf Experiments compute resources}
    \item[] Question: For each experiment, does the paper provide sufficient information on the computer resources (type of compute workers, memory, time of execution) needed to reproduce the experiments?
    \item[] Answer: \answerYes{} % Replace by \answerYes{}, \answerNo{}, or \answerNA{}.
    \item[] Justification: Appendix~\ref{app:compute} reports the hardware and the GPU-hour cost of caption generation and of each audit stage.
    
\item {\bf Code of ethics}
    \item[] Question: Does the research conducted in the paper conform, in every respect, with the NeurIPS Code of Ethics \url{https://neurips.cc/public/EthicsGuidelines}?
    \item[] Answer: \answerYes{} % Replace by \answerYes{}, \answerNo{}, or \answerNA{}.
    \item[] Justification: Yes, all authors follow NeurIPS Code of Ethics.

\item {\bf Broader impacts}
    \item[] Question: Does the paper discuss both potential positive societal impacts and negative societal impacts of the work performed?
    \item[] Answer: \answerYes{} % Replace by \answerYes{}, \answerNo{}, or \answerNA{}.
    \item[] Justification: The audit makes caption supervision inspectable before a generator is trained (Section~\ref{sec:intro}); Section~\ref{sec:release} and Appendix~\ref{app:release_manifest} cover personal information in generated captions, the filtering that training use requires, and the removal channel.
    
\item {\bf Safeguards}
    \item[] Question: Does the paper describe safeguards that have been put in place for responsible release of data or models that have a high risk for misuse (e.g., pre-trained language models, image generators, or scraped datasets)?
    \item[] Answer: \answerYes{} % Replace by \answerYes{}, \answerNo{}, or \answerNA{}.
    \item[] Justification: The release is caption-only; every dataset card states that captions are released as generated, lists the filtering required before training use, and gives a removal channel (Appendix~\ref{app:release_manifest}).

\item {\bf Licenses for existing assets}
    \item[] Question: Are the creators or original owners of assets (e.g., code, data, models), used in the paper, properly credited and are the license and terms of use explicitly mentioned and properly respected?
    \item[] Answer: \answerYes{} % Replace by \answerYes{}, \answerNo{}, or \answerNA{}.
    \item[] Justification: Yes, we respect dataset's license and mention explicitly.

\item {\bf New assets}
    \item[] Question: Are new assets introduced in the paper well documented and is the documentation provided alongside the assets?
    \item[] Answer: \answerYes{} % Replace by \answerYes{}, \answerNo{}, or \answerNA{}.
    \item[] Justification: Appendix~\ref{app:release_manifest} documents the public assets, and each dataset carries a dataset card and a Croissant record.

\item {\bf Crowdsourcing and research with human subjects}
    \item[] Question: For crowdsourcing experiments and research with human subjects, does the paper include the full text of instructions given to participants and screenshots, if applicable, as well as details about compensation (if any)? 
    \item[] Answer: \answerYes{} % Replace by \answerYes{}, \answerNo{}, or \answerNA{}.
    \item[] Justification: Seven volunteer annotators verified machine-extracted
    claims against captions and images without compensation;
    Appendix~\ref{app:human_cbu_details} shows the annotation interface, and the
    code repository carries the complete instruction sequence.

\item {\bf Institutional review board (IRB) approvals or equivalent for research with human subjects}
    \item[] Question: Does the paper describe potential risks incurred by study participants, whether such risks were disclosed to the subjects, and whether Institutional Review Board (IRB) approvals (or an equivalent approval/review based on the requirements of your country or institution) were obtained?
    \item[] Answer: \answerNA{} % Replace by \answerYes{}, \answerNo{}, or \answerNA{}.
    \item[] Justification: The human input is a verification pass over
    machine-extracted claims (Appendix~\ref{app:human_cbu_details}); the task records
    judgments about captions and images and collects no data about the annotators.

\item {\bf Declaration of LLM usage}
    \item[] Question: Does the paper describe the usage of LLMs if it is an important, original, or non-standard component of the core methods in this research? Note that if the LLM is used only for writing, editing, or formatting purposes and does \emph{not} impact the core methodology, scientific rigor, or originality of the research, declaration is not required.
    %this research? 
    \item[] Answer: \answerYes{} % Replace by \answerYes{}, \answerNo{}, or \answerNA{}.
    \item[] Justification: VLMs serve as the captioner, the CBU extractor, and the image-conditioned VQA Judges.

\end{enumerate}

\end{document}